\documentclass[journal]{IEEEtran}

\usepackage[usenames, dvipsnames]{color}
\usepackage{amsmath}
\usepackage{amssymb}
\usepackage{cite}
\usepackage{graphicx} 
\usepackage{multicol}
\usepackage{epstopdf}
\usepackage{verbatim}
\usepackage{url}
\usepackage{amsmath}
\usepackage{breqn}
\usepackage{amsthm}
\usepackage{multirow}
\usepackage{lipsum}
\usepackage{bbm}
\usepackage{caption}
\usepackage{algorithm}
\usepackage{mathtools}

\newtheorem{thm}{Theorem}
\newtheorem{lemma}{Lemma}
\newtheorem{prop}{Proposition}

\newtheorem{defn}{Definition}

\usepackage{dblfloatfix}
\usepackage{array}

\usepackage{enumitem}
\renewcommand{\baselinestretch}{.9}

\usepackage{microtype}
\usepackage{subfigure}
\usepackage{booktabs} 

\usepackage{algorithmic}

\usepackage{hyperref}

\begin{document}

\title{Multi-Player Bandits: A Trekking Approach}

\author{Manjesh K.~Hanawal
	and Sumit J.~Darak
	\IEEEcompsocitemizethanks{\IEEEcompsocthanksitem Manjesh K. Hanawal is with IEOR, IIT Bombay, India. E-mail: mhanawal@iitb.ac.in. 	
			
		Sumit J. Darak is with the Department of ECE, Indraprastha Institute of Information Technology, Delhi, India. E-mail: sumit@iiitd.ac.in. 
}

}
\maketitle

\begin{abstract}
We study stochastic multi-armed bandits with many players. The players do not know the number of players, cannot communicate with each other and if multiple players select a common arm they collide and none of them receive any reward. We consider the static scenario, where the number of players remains fixed, and the dynamic scenario, where the players enter and leave at any time. We provide algorithms based on a novel `trekking approach' that guarantees constant regret for the static case and sub-linear regret for the dynamic case with high probability. The trekking approach eliminates the need to estimate the number of players resulting in fewer collisions and improved regret performance compared to the state-of-the-art algorithms.  We also develop an epoch-less algorithm that eliminates any requirement of time synchronization across the players provided each player can detect the presence of other players on an arm. We validate our theoretical guarantees using simulation based and real test-bed based experiments.
\end{abstract}


\IEEEpeerreviewmaketitle

\section{Introduction}
\label{sec:submission}

Multi-player multi-armed bandits (MPMAB) is a variant of the stochastic multi-armed bandits \cite{mab1,mab2,mab3} where multiple players aim to maximize sum of their rewards playing the same set of arms. In this setting, the players do not communicate with each other and may not know number of other players in the game. If two or more players select the same arm simultaneously, they experience `collision'  and none of them receive any reward. Our goal in this work is to develop distributed algorithms that aim to achieve high total rewards while keeping the number of collisions as low as possible.

The study of MPMAB is mainly motivated from the ad-hoc cognitive radio networks (CRN) where multiple users transmit on a common set of channels (unlicensed spectrum) without any communication among them \cite{ad_hoc_1,ad_hoc_2}. Due to the ad hoc nature of such networks, a central controller, or a common control channels for coordination, may not be available and all channel selection decisions have to be done in a decentralized fashion \cite{ad_hoc_1,ad_hoc_2,ad_hoc_0}. Such models are being envisioned for futuristic ultra-dense wireless communication networks that can offer very high peak rates \cite{NR_1}. The quality of the channels are unknown to the users and their goal is to maximize number of successful transmissions (or sum rate/ throughput) in the network. In a CRN the users not only have to learn the channel qualities but also have to learn to co-ordinate by selecting non-overlapping channels. The MPMAB provides the required `learning-to-coordinate' framework in a distributed fashion.

At each round of the game, each player selects an arm to play. If an arm is played by only one player, that player receives reward associated with the arm, otherwise, all the players selecting it observe collision and receive zero reward. Further, a player experiencing a collision will not know with whom and how many she collided. The performance of a policy is measured as the difference of the total expected reward from it and that from the policy that selects non-overlapping arms form the top $N$ arms in each round for all the players. Here the top $N$ arms refer to the set of $N$ arms with highest mean rewards. The total number of collision is the sum of collisions experienced by all the players.

For applications like CRN where the players are mostly battery operated, higher number of collisions results in reduced operational life. Hence it is desirable that the algorithms for MPMAB should work with as fewer number of collisions as possible. The state-of-the-art Musical Chair (MC) \cite{ICML16_MultiplayerBandits_RosenkiShamir} algorithm forces a certain number of collisions in the game to estimate the number of players even though its regret performance is superior compared to other algorithms (for unknown number of players).  In this work, we develop algorithms based on novel `trekking approach' that significantly reduces the number of collisions in the game while guaranteeing a good regret performance.

The trekking approach is based on the simple idea that once the players have a good estimate of arms they should try to pick their next-best arm in their ordered-list till collision is experienced on an arm. Once a collision is observed, they should return (after some back-off time) to the arm on which previously no collision was observed and play it till the end. We refer to this process of continuously looking for the next-best arm as `trekking'. When the process ends, this approach ensures that all the players are playing the top $N$ arms. 

As in \cite{ICML16_MultiplayerBandits_RosenkiShamir} and \cite{KDD14_ConcurrentBandits_AvnerMannor}, we consider two variants of multi-player bandits-- static and dynamic. In the static case, all the players start the game simultaneously and continue till the end. In the dynamic case, the players can enter and leave the game at any point. For both the cases, we provide trekking based algorithms with high confidence bounds on regret and collisions. Similar to MC, our algorithm for the dynamic case needs to restart after a certain number of rounds. To overcome this limitation, we propose an epoch-less algorithm that works provided the players can check if any other player is playing the arm they selected. We refer to this requirement as `sensing capability'. This requirement is readily satisfied in CRNs where each player is equipped with capabilities that enable them to check if any other player is transmitting on the channel they like to use. Note that a player can either transmit or sense but not both. Our main results assuming a fixed gap between the mean rewards are as follows:
\begin{itemize}
	\item For the static case we propose and analyze Static Trekking (ST) algorithm that guarantees constant regret and collisions (independent of number of rounds) with high probability (w.h.p). 
	\item For the dynamic case we propose and analyze Dynamic Trekking (DT) algorithm that guarantees $\mathcal{O}(\sqrt{xT})$ regret and collisions w.h.p, where $T$ is the time horizon and $x$ is the bound on the number of players entering and leaving the game. This algorithm restarts at regular periods (epochs) that depends on $T$. 
	\item We show that the regret and collisions in our algorithms is lower by a factor of $4$ and $12.5K$, respectively, than the state-of-the art algorithms.
%
		\item When players have sensing capability, we propose and analyze Dynamic Trekking with Sensing (DTS)  algorithm which does not require to restart (epoch-free) but guarantees
		 $\mathcal{O}(\sqrt{xT})$ regret and collisions w.h.p. DTS does not require any time synchronization of players.
			\item Finally, we validate the theoretical guarantees through experiments based on synthetic and real test-bed setup.  Both regret and collisions are lower in our algorithms compared to the state-of-the-art. Source code of all implementations is available online at  \cite{code}.
\end{itemize} 
\subsection{Related Work}
Most works on stochastic bandits with multiple-payers require some negotiation or pre-agreement phase to avoid collisions between the players. The $\mbox{dUCB}_4$ algorithm in  \cite{TIT14_DecentralizedLearning_KalthilNayyarJain} achieves this using Bertsekas' auction mechanism for players to negotiate unique arm. The Time Divisions Fair Sharing (TDFS) algorithm in \cite{TSP10_DistributedLearning_LiuZhao} requires players to agree on a time division of slots before the game. Such negotiations are hard to realize in a completely distributed setup like ad hoc CRN \cite{ad_hoc_1,ad_hoc_2}.  The $\rho^{\mbox{{\tiny RAND}}}$ algorithm in \cite{JSAC11_DistributedLearning_Anadakumar} is communication free and completely decentralized. Performance improvements of  $\rho^{\mbox{{\tiny RAND}}}$ are studied recently in \cite{ALT2018_MultiplayerBandits_BessonKaufma}. However, these algorithms consider only static case and assumes prior knowledge of number of players. The modified $\rho^{\mbox{{\tiny EST}}}$ algorithm overcomes latter issue, but its guarantees holds only asymptotically. Other set of works in \cite{TWC16_DistributedStochastic_ZandiDongGrami}, \cite{TMC11_CongnitiveMediumAccess_LaiGamalJiangPoor} considers selfish behavior of players and analyze their equilibrium behavior. However, all these algorithms work only for the static case and cannot extend to the dynamic scenarios which is the focus of this work.


The works most similar to ours are \cite{INFOCOM16_MultiUserLax_AvnerMannor}, \cite{KDD14_ConcurrentBandits_AvnerMannor} and \cite{ICML16_MultiplayerBandits_RosenkiShamir} which consider communication free setting with unknown number of players that can vary during the game. The algorithm in \cite{INFOCOM16_MultiUserLax_AvnerMannor} also considers the case where the arm characteristics are different across players but does not guarantee network optimal reward. The major drawback of this algorithm is that it assumes that the players gets to know information of the channels selected by all other players in each time slot. In ad-hoc CRN, complex hardware is needed to gain such information and hence, it is not feasible for battery operated users \cite{sns}. The MEGA algorithm in \cite{KDD14_ConcurrentBandits_AvnerMannor} uses the classical $\epsilon$-greedy MAB algorithm and ALOHA based collision avoidance mechanism. Though collision frequency reduces in MEGA as the game proceeds it may not go to zero as shown in \cite{ICML16_MultiplayerBandits_RosenkiShamir}. To overcome this \cite{ICML16_MultiplayerBandits_RosenkiShamir} develop Musical Chairs (MC) algorithm that incurs collisions only in the initial phase and guarantees collision free play subsequently. Though MC performs better than MEGA, its performance in the initial rounds is poor -- MC uses collision information to estimate the number of players and forces a large number of collisions to get a good estimate. 

Our approach reduces total collisions  by circumventing the need to estimate the number of players and guarantees collision free play after few rounds.
The first part of our algorithms find orthogonal arm allocations through random hopping and then follow a common (deterministic) hopping pattern which is similar to the two phase channel/arm access scheme in \cite{JSAC2014_DesignAnalysis_ZhangYao}. However, it considers only the static case with identical arms.

The trekking approach has been discussed in \cite{wiopt} where we considered the static case. In this paper, we provide algorithms for both the static and dynamic case and their analysis. The proposed algorithms in \cite{wiopt} are specifically designed for CRN in a licensed spectrum whereas the current work focuses on a more general MPMAB setting and hence the analysis is substantially different than in \cite{wiopt}.

\noindent
{\bf Organization of the paper:}In Section \ref{sec:setup} we introduce the notations and setup the problem. In Section \ref{sec:AlgoNoSensing} we give algorithms and analyze their performance for the static and dynamic scenarios in sections \ref{sec:AlgoNoSensingd} and \ref{sec:AlgoNoSensingd}, respectively.  In section  \ref{sec:AlgoSensing}, we modify the algorithm for the dynamic scenario to work without requiring any shared global clock.  We validate our claims through both synthetic and real test-bed setup in Section \ref{sec:SR}. Conclusions and future directions are given in Section \ref{sec:Conclu}. All the proofs are in the appendix given at the end of the paper.


\section{ Problem Setup}
\label{sec:setup}

The standard stochastic $K$-armed bandit consists of a single player with $K>1$ arms. Playing arm $k \in [K]$ gives reward drawn independently from a distribution with support $[0 \; 1]$.   The reward distributions are stationary and independent across the players. Let $\mu_k$ denotes the mean of arm $k$ and $\mu^*=\max_{k}\mu_k$ is the largest mean. 
The multi-player $K$-armed bandit is similar, but consists of multiple players that can vary with time. Let $N_t \leq K$\footnote{For ease of analysis, we assume $N_t \leq K$ i.e. the number of players are less than the number of arms. However, later we show that the proposed algorithms can work when  $N_t > K$.} denotes the number of players in round $t$. The players are not aware of how many other players are present and cannot communicate with each other. We consider the static case where $N_t=N$ for all $t$ and the dynamic case where $N_t$ can change with $t$.  When a player selects an arm, reward is obtained if only she happens to play that arm, otherwise all the players choosing that arm will get zero reward. We refer to the latter case as 'collision'. For any distributed policy in which player $k$ plays arm $I_{k,t}$ in round $t$, expected regret over period $T$ is defined as 
\begin{equation}
R=\sum_{t=1}^T \sum_{k \in K_t^*} \mu_{k} -\sum_{t=1}^T\sum_{j \in [N_t]}\mu_{I_{j,t}}(1-\eta_{j,t})
\end{equation}  
where $K_t^*$ denotes the set of arms with $N_t$ highest mean rewards, i.e., the set of top $N_t$ arms. $\eta_{j,t}$ is collision indicator for player $k$ in round $t$. It is set to $1$ if more than one player select arm $j$ in round $t$, otherwise it is set to $0$.
The total number of collision over period $T$ is defined as
\begin{equation}
C=\sum_{t=1}^T \sum_{j \in [N_t]}  \eta_{j,t}
\end{equation}
Our goal is to develop distributed algorithms that minimizes $R$ while keeping $C$ as low as possible.

\section{Static Trekking Algorithm}
\label{sec:AlgoNoSensing}
We first consider the static case where the number of players remains fixed throughout the game and develop an algorithm named Static Trekking (ST) based on the novel trekking approach. 
\subsection{ST Algorithm}
The ST algorithm works in two phases namely, learning phase and trekking phase. In the learning phase, each player initially plays an arm drawn uniformly at random in each round. Once a player observes collision-free play on an arm, she starts playing the arms sequentially drawing an arm with higher index (up to modulo $K$) in each round. After all the players observe a collision-free play, the arms played by them are orthogonal in each round and no collisions occur. We refer to the part of learning phase in which all players orthogonalize as Random Hopping (RH) sub-phase and the part in which each  player select arms sequentially as Sequential Hopping (SH) sub-phase. The learning phase runs for $T_0$ rounds which is set such that all players find orthogonal arms and learn correct ranking of the arms with high probability. After $T_0$ rounds, each player re-index the arms according to decreasing value of their estimated means.  If a player is on the top channel in the $T_0$ round she continues to play it henceforth, otherwise, she enters into the trekking phase. 

In the trekking phase, each player sets the arm played at the end of learning phase as their reserved arm and checks for availability of their next best arm. Specifically, a player updates its current reserved arm, say $i$,  to $i-1$ if no collision is observed on arm $i-1$ in the next $i-1$ rounds. Otherwise she goes back to arm $i$ and plays it in all the subsequent rounds. We refer to this latter scenario as `player is locked'. When a player's reserved arm changes, her previously reserved arm is `released' and can become a reserved arm for another player. Updating of a reserved arm is continued either till the player is locked, or the top arm becomes her reserved arm in which case she locks on it. Thus, in the trekking phase, the players `trek' towards the better arms and all $N$ players settle on one of the distinct top $N$ arms. 
\begin{algorithm}[!t]

	\caption{Static Trekking (ST)} \label{Algo1}
	\begin{algorithmic}
		\STATE	Input: {$K, \delta$ }
		\STATE Compute $T_0$ as in (\ref{eqn:ST_T0})
		\STATE	$(\hat{\mu}, I)$= Learning $(T_0, K)$
		\STATE	 Trekking$( \hat{\mu}, I)$ 
	\end{algorithmic}
\end{algorithm}

\begin{algorithm}[!t]
	\caption*{\textbf{Subroutine:} Learning}
	\begin{algorithmic}[1]
		\STATE Input: $T_{0}, K$ 
		\STATE Set $L=0$ and $V_k=0, S_k=0 \;\; \forall k \in [K]$ 
		\FOR{$t=1 \dots T_{0}$}
		\IF{($L == 1$)}
		\STATE Choose arm, $I_t = I_{t-1}+1$ modulo $K$
		\ELSE
		\STATE Randomly choose arm, $I_t \sim U(1, ... ,K)$ 
		\ENDIF
		\STATE Increment $S_{I_t}$ by 1
		\IF{ no collision ($\eta_{I_t,t}==0$)}
		\STATE Set $L=1$ and $V_{I_t}\leftarrow V_{I_t} + r_{I_t}$
		\ENDIF
		\ENDFOR	
		\STATE Estimate arm means, $\hat{\mu}_k = \frac{V_{k}}{S_{k}} \; \forall k$. $\hat{\mu}:=\{\hat{\mu}_k\}_{k \in [K]}$
		\STATE Re-index arms according to their rank in $\hat{\mu}$. Call it $\pi$
		\IF {Index of $I_{T_0}$ in $\pi$ is $1$}
		\STATE Play first arm in $\pi$ henceforth
		\ELSE
		\STATE Trekking($\pi$, Index of $I_{T_0}$ in $\pi$)
		\ENDIF      
	\end{algorithmic}
\end{algorithm}
\begin{algorithm}[!t]
	\caption*{\textbf{Subroutine:} Trekking (TrekU)}
	\begin{algorithmic}[1]
		\STATE Input : {$\pi$ (ordered list of arms), $J$ (arm index) } 
		\STATE Set $Y_k= 1\;\; \forall k \in [K]$ and  $L=0$ (lock indicator) 
		\FOR {$t=T_{0}+1 \dots $}
		\IF{$L==1$}
		\STATE Select the same arm, $I_t=I_{t-1}$
		\ELSIF{$Y_{J}\leq J-1$}
		\STATE Select the (same) next best arm, $I_t=J-1$
		\STATE  $Y_{J}\leftarrow Y_{J} + 1$
		\ELSIF {$I_{t-1}$ not equals to $1$}
		\STATE Select the next best arm, $I_t=I_{t-1}-1$ 
		\STATE Update reserved arm $J=I_{t-1}$
		\ELSE 
		\STATE Lock on the top arm, $L=1$ 
		\ENDIF
		\IF{ collision ($\eta_{I_t,t}==1$)}
		\STATE Lock on reserved arm, $I_{t}=J$ and $L=1$	
		\ENDIF
		\ENDFOR
	\end{algorithmic}
\end{algorithm}	


Observing arm $i$ for $i$ rounds by a player before making it as her reserved arm prevents two players from locking on the same arm. To see this, consider that arm $2$ is the reserved arm for a player. This player needs one slot to check if arm $1$ is taken by any other player (by observing collision on it), and in case it is taken, she needs another slot to return and get locked on arm $2$. During these $2$ slots the player with reserved arm $3$ should not lock on arm $2$. Extending this argument for any arm $i$, if each player observes arm $i$ for $i$ rounds, all players are ensured not to lock on the same arm and orthogonality is achieved on the top $N$ arms. The pseudo-code of ST is given in  Algorithm \ref{Algo1} which is run by each player faithfully. We suppress the player index to simplify notations. 

The success of trekking phase depends on the players having the correct ranking of the arms. We next show that setting $T_0$ long enough, each player learns correct ranking of arms and the trekking phase then guarantees that each player will lock on one of the top $N$ distinct arm within a bounded number of rounds and no regret is incurred after that. 
\subsection{Analysis of ST Algorithm}
\label{sec:AST}
In this subsection, we bound the expected regret and number of collisions of the ST algorithm. For each player let $\hat{\mu}_k$ denotes the empirical mean of arm $k$. We begin with the following definition given in \cite{ICML16_MultiplayerBandits_RosenkiShamir}.

\begin{defn}
	An $\epsilon$-correct ranking of $K$ arms is a sorted list of their empirical mean such that $\forall i, j: \hat{\mu}_i$ is listed before $\hat{\mu}_j$ if $\mu_i - \mu_j \geq \epsilon$.
\end{defn} 

\begin{thm}
	\label{thm:StaticST}
Let $\Delta>0$ be the gap between the mean rewards of $N^{th}$ and $(N+1)^{th}$ best arm. Then for all $\epsilon < \Delta$ and $\delta \in (0,1)$, with probability at least $1-\delta$, the expected regret of the ST algorithm from $T$ rounds is bounded as
\begin{equation}
\label{eqn:STRegret}
R \leq N(T_{rh} + T_{sh}(1-N/K) + T_{tr}),
\end{equation}
where $T_{sh}:=T_{sh}(\delta/2), T_{rh}:=T_{rh}(\delta/2)$ and $T_{tr}$ are given as follows:  
\begin{equation*}
\label{eqn:ST_TL}
\hspace{-.2cm}T_{rh}=\left \lceil \hspace{-.1cm}\frac{\log\left (\frac{\delta}{2K} \right)} {\log\left(1-\frac{1}{4K}\right)} \hspace{-.1cm}\right\rceil,
\end{equation*}
\begin{equation*}
 T_{sh}=\frac{2K}{\epsilon^2}\log \left (\frac{4KN}{\delta}\right),
\end{equation*}
\begin{equation*}
T_{tr}=(K^2-(N-1)^2)/2 + 1.
\end{equation*}

Further, the number of collisions is bounded  with probability at least $1-\delta$ as
\begin{equation}
C\leq N T_{rh} + 4N.
\end{equation}

\end{thm}
A total of $T_0=T_{rh}+T_{sh}$ rounds guarantee that all players orthogonalize and learn $\epsilon$-correct ranking of the arms with probability at least $1-\delta$. The length of the learning phase is set to $T_0$ in ST algorithm.  Note that the value of $T_{sh}$ depends on $N$, which may not be known. We redefine $T_{sh} $ by replacing $N$ with $K$ and use the following value of $T_0$ which upper bounds earlier value and depends only on $K$ and $\delta$.
%

 \begin{equation}
 \label{eqn:ST_T0}
T_0=\left \lceil \frac{\log(\delta/2K)}{\log\left(1-1/4K\right)}\right \rceil + \frac{2K}{\epsilon^2}\log \left (\frac{4K^2}{\delta}\right).
 \end{equation}

The bounds hold under the assumption that a lower bound on the reward gap is known. This assumption is also made in \cite{KDD14_ConcurrentBandits_AvnerMannor} and \cite{ICML16_MultiplayerBandits_RosenkiShamir}.  The bounds are in expectation and conditioned on the fact that that players learn the $\epsilon$-correct ranking of the arms which happens with probability $1-\delta$ if learning phase is run for $T_0$ number of rounds.

The proof of Thm \ref{thm:StaticST} is given in the Appendix. The proof consists of bounding  the expected number of rounds required for each player to 1) observe a collision-free play through random selection of arms 2) learn $\epsilon$-correct ranking of arms and 3) settle on one of the top $N$ arms through trekking.  All the bounds are independent of $T$. After the three events, each player will lock on one of the top $N$ distinct arm hence no regret in the subsequent plays. The collision bound is obtained by bounding the expected number of rounds for all the players to orthogonalize in the learning phase and noting that at most two collisions are observed by each player during the trekking phase. 

In contrast to the MC algorithm, the ST algorithm aims to orthogonalize the players as early as possible in the learning phase. This significantly brings down the collisions and improves the regret. We next compare the collision and regret performance of the state-of-the-art MC algorithm.
\subsection{Performance comparison with the MC algorithm}\label{staticthm}
 The MC algorithm also works in phases. In the learning phase, each player selects an arm uniformly at random from $[K]$ in each round and obtains $\epsilon$-correct ranking of the arms and an estimate of number of players. In the next phase, named musical chairs, each player plays an arm selected uniformly at random from the top $N$ arms till it observes a collision-free play on an arm and locks on it. 
 \begin{thm}
 	\label{thm:MCBound}
 For the same setup in Thm \ref{thm:StaticST}, the expected regret of MC with probability at least $1-\delta$ is bounded as 
 \begin{equation}
 \label{eqn:RegretMC}
 R^{MC}\leq NT_0^{MC} + 2N^2\exp(2),
 \end{equation}
 where
 \begin{equation*}
 T_0^{MC}= \max\left( \frac{16K}{\epsilon^2}\ln \left(\frac{4K^2}{\delta}\right), \frac{K^2\log (4/\delta)}{0.02} \right).
 \end{equation*}

The expected number of collisions in the MC algorithm with $N\geq 2$ is lower bounded as 
 \begin{equation} 
 C^{MC}>(N/K)T_0^{MC}.
  \end{equation}
\end{thm}
The proof for regret is given in \cite{ICML16_MultiplayerBandits_RosenkiShamir} and the proof for collision is given in the Appendix. 

{\bf Regret comparison:} The dominant terms in the regret bounds of the ST and MC algorithms are  $\frac{2K}{\epsilon^2}\log \left (\frac{4K^2}{\delta}\right)$
(Eq. (\ref{eqn:STRegret}), (\ref{eqn:ST_T0})) and $\max\left( \frac{16K}{\epsilon^2}\ln \left(\frac{4K^2}{\delta}\right), \frac{K^2\log (4/\delta)}{0.02} \right)$ (Eq. \ref{eqn:RegretMC}), respectively. Comparing the two, the regret bound of ST is smaller by at least a factor of $8$. MC obtains ranking of arms by uniform sampling of arms which requires about $4$ plays of an arm to get one reward sample. Whereas in ST reward is observed in each round after orthogonalization. Thus giving a gain of factor $4$. Another factor $2$ in MC is due to an application of Chernoff bound (this can be tightened). 

{\bf Collision comparison:} The average number of collisions incurred by ST during the learning phase is no more than that  incurred by MC during the MC phase -- in the former case an arm is selected uniformly at random from set $[K]$ whereas it is  from the small set $[N]$ in the latter case till a collision-free arm is found. We next compare the number of collisions in the trekking phase of ST and learning phase of MC. In the former case collisions are at most $4K$, whereas it is at least $\frac{K^2\log (4/\delta)}{0.02}$ in the latter case. Hence collisions in ST are smaller by at least a factor of $12.5K$ compared to the MC algorithm.

When $N=K$, the regret in ST is minimal -- regret is non-zero only in the $T_{rh}$ rounds of the learning and $K$ rounds of the trekking phase. Whereas the performance of  MC degrades as $N$ increases. In Section \ref{sec:SR}, Fig \ref{Algo1} we see that the difference in regret of ST and that of MC increases with $N$ validating the proposed hypothesis. 

\begin{algorithm}[!b]
	\caption*{\textbf{Subroutine:} Modified Trekking (TrekD)}
		\label{alg:TrekS1}
	\begin{algorithmic}[1]
		\STATE Input : \hspace{-.1cm}{$\pi_1$ (ordered list of estimated arms), $J$ (arm index) } 
		\STATE Set $Y_k= 1\;\; \forall k \in [K], L=0, R=1, J_n=1$  
		\FOR {$t=T_{0}+1 \dots $}
		\IF{locked ($L==1) $}
		\STATE Select the same arm, $I_t=J$
		\ELSIF { returned to reserved arm ($R==1$)}
		\STATE Select the next `worst' arm $I_t \leftarrow \pi_1(J_n)$ and sense
		\STATE Update next `worst' arm $J_n \leftarrow J_n +1$, set $R \leftarrow 0$ and $Y_{I_t} \leftarrow Y_{I_t} +1$
		\ELSIF{collision ($\eta_{I_{t-1},{t-1}}==1$)}
		\STATE Set $Y_{I_t} \leftarrow Y_{I_t} +1$
		\IF {back-off time lapsed ($Y_{I_t}> K-J$)}
		\STATE Return to reserved arm $R\leftarrow 1$ 
		\ELSE
		\STATE Stay on the same arm $I_t \leftarrow I_{t-1}$
		\ENDIF
		\ELSE 
		\STATE	Set $L\leftarrow 1, I_t\leftarrow I_{t-1},  J\leftarrow I_{t-1}$
		\ENDIF
		\ENDFOR
	\end{algorithmic}
\end{algorithm}		

\subsection{ST Algorithm Using Modified Trekking Approach}
\label{sec:mtrek}
In this subsection, we discuss another version of the ST algorithm that is more suitable to the dynamic version where players can enter and leave anytime. In the current version if a player on the top arm leaves, regret is incurred till all the players shift to their next best arm. But, if players start from the top arm and go down as per the ordered list (trekking downwards) instead of checking for their next best arm (trekking upwards), any arm freed up by a leaving player will be taken up earlier. However, in the new 'trekking downward' approach many players may select an arm simultaneously. To overcome this, we introduce 'back off' mechanism to resolve which player locks on an arm.  The ST algorithm based on these modifications is described below and its pseudo-code is given below. We refer to the earlier trekking approach as TrekU and the modified version as TrekD.

Suppose that a player is on arm $i$ at the end of learning phase. The player sets it back-off time as  $K-i+1$ rounds and it remains same for entire trekking phase. A player begins trekking by playing an arm $1$ for the next $K-i+1$ rounds and set the arm $2$ as its reserved arm. If a collision free play is observed within the $K-i+1$ rounds, she enters into lock state and updates her reserved arm to $1$. We again refer to this scenario as `player is locked'. If collisions are observed in each of the $K-i+1$ rounds, the player moves to the reserved arm, i.e. $2$, and updates the next reserved arm as $3$.  The player then check for availability for arm $2$ applying the same procedure. The process is repeated until player gets locked on one of the top arms. Thus, in the modified trekking phase, the players `trek' from the top arm towards bottom arm and all $N$ players settle on one of the distinct top $N$ arms. 

The number of rounds required by TrekD to settle the players on the top $N$ channels is at most $(N-1)(K-1)+1$ (see Lemma~\ref{lma:Trekking_Static1} in the Appendix). This is higher than the corresponding $(K^2 -(N-1)^2)/2+1$ bound for the TrekU. It is not hard to realize scenarios where the bounds are tight for both cases. Though TrekU is better on an average, we will see later that TrekD extend to the dynamic scenarios more naturally.



\section{Epoch Based Dynamic Trekking Algorithm}
\label{sec:AlgoNoSensingd}
Here, we consider the dynamic case where the number players can enter and leave the game anytime. The proposed algorithm, named Dynamic Trekking (DT) algorithm, runs in epochs and restarts after each epoch. The rate at which epochs restarts is set based on duration of the game $T$. The algorithm requires that all the players restart the epochs at the same time. Such requirement can be achieved through a global clock as discussed in \cite{ICML16_MultiplayerBandits_RosenkiShamir}. The pseudo-code of the DT algorithm is given in \ref{alg:Algo2} where $t$ denotes the time on the global clock and $T_1$ is the length of each epoch. The DT algorithm can work using any one of the trekking approach.
\begin{algorithm}[!h]
	\caption{Dynamic Trekking (DT)} \label{alg:Algo2}
	\begin{algorithmic}
		\STATE	Input: {$K, \delta, T,x$} 
		\STATE Compute $T_0$ as in (\ref{eqn:ST_T0}) and $T_{ep}$ as in (\ref{eqn:EpochLen})
		\IF {$ t \mod T_{ep} ==0$  } 
		\STATE	$(\hat{\mu}, I)$= Learning $(T_0, K)$
		\STATE	 Trekking $( \hat{\mu}, I)$ 
		\ENDIF
	\end{algorithmic}
\end{algorithm}

The performance guarantee of the DT algorithm is provided under the following additional assumptions : 1) number of players entering and leaving is bounded or at most sub-linear in $T$, 2) no new player enters during the learning and trekking phase in each epoch. If number of active players changes frequently, no learning may be possible and regret is linear. The first assumption restricts this behavior. The second assumption ensures that the players who joined at the beginning of an epoch get correct ranking of arms. 

\begin{prop}
\label{prop:DynamicDT}
Let at most $x$ players enter and leave during $T$ rounds, and $\Delta_m=\min_{ i\in [K-1]} \mu_i - \mu_{i+1}$. Then for all $\delta \in (0,1)$ and $\epsilon < \Delta_m$ with probability at least $1-\delta$ the expected regret of the DT algorithm after $T$ rounds with $T_0:=T_0(\delta/T)$ is
\begin{equation*}
R < TK(T_0 + T_{tr})/T_{ep} + 2x(T_{ep}-T_0-T_{tr}),
\end{equation*}
\end{prop}
and the expected number of collisions is
\begin{equation*}
C \leq (T/T_{ep})\left ( KT_{rh}(\delta/2T) + 4K \right) + 2x(T_{ep}-T_0-T_{tr}),
\end{equation*}
where $T_{tr}$ is a constant and $T_{ep}$ grows sub-linearly given by
\begin{equation}
\label{eqn:EpochLen}
 T_{ep}=\sqrt{\frac{TK(T_0+T_{tr})}{2x}}.
\end{equation}

\begin{thm}
\label{thm:DynamicDT}
	For the setup in Prop. \ref{thm:DynamicDT} with high probability the expected regret and collisions in DT with length of the epoch period set to $T_{ep}$ and learning phase set to $T_0$ are, respectively,
	\[R\leq \tilde{O}(\sqrt{xT}) \mbox{  and  } C\leq \tilde{O}(\sqrt{xT}),\]
	where the $\tilde{O}$ hides logarithmic factors.
\end{thm}

For each epoch, the proof bounds regret for the players who are present from the start of the epoch. The bound is obtained exactly as in static case but by setting the confidence interval to $\delta/T$ which makes the length of learning period to be logarithmic in $T$.  This regret is aggregated over all epochs and a high confidence is obtained after applying union bound. The regret due to entering and leaving users is bounded separately which depends on the length of the epochs. The final bound is obtained by summing regret from all type of players and optimizing over epoch length. We require $T_{ep}$ to be larger than $T_0$, if not we set $T_{ep}=T_0$.

The assumptions made here for dynamic setting are also used in \cite{ICML16_MultiplayerBandits_RosenkiShamir} where it is further assumed that players do not leave during the learning phase. We allow the players to leave at any time. This is possible because  trekking approach ensures all the players settle on the top arms without knowing how many are present. Whereas this is not possible in the MC algorithm -- if players leave during the learning phase, the estimate of number of players can be incorrect and MC can fail. Further, the total number of rounds for players to learn and lock in each epoch in DT is smaller (by a factor of at least $4$) than length of the learning phase in DMC leading to least a factor of $4$ improvement in regret and $12.5K$ in collisions (as in ST). Also, DT works with fewer restrictions than DMC. 

{\bf Remark:} The trekking approach allows to handle the case $N>K$, i.e., more number of players than arms. In ST once a player gets a collision-free play through RH sub-phase, it switches to SH sub-phase. Because of this some players will observe continuous collisions within the RH sub-phase if $N>K$ and can leave in at most $T_{rh}$ time slots. In DMC, players are always in RH sub-phase and it is possible that no one will observe continuous collisions. Then, it is unclear after how many collisions they should leave and which one of them should leave.
Another major issue in MC is that it will fail if players leave in between. This is because MC uses collision count to estimate $N$. If players leave, collision count will not give correct estimate of $N$. In trekking approach, players need not know $N$ and works even if $N$ changes.


The limitation of DT and DMC is that they require a global clock so that all the players can restart their epochs simultaneously. But, in a completely decentralized systems, like  CRN, this may not be possible. However, what is possible in applications like CRN is that players can check/detect presence of other players on the arms before playing them. We exploits this ability and propose an epoch-less algorithm that relaxes the need to have a global clock for synchronization.

\section{Epoch-Free Dynamic Trekking Algorithm}
\label{sec:AlgoSensing}
In the dynamic case, a player entering late can disturb the trekking process of other players and prevent them from locking on top arms. This can be avoided if a new entrant plays an arm only if no locked player is detected on that arm. This feature can be readily available in applications like CRN where each player is equipped with a transmitter and receiver pair -- transmitter sends information on a channel/arm while receiver detects collision \cite{JSAC11_DistributedLearning_Anadakumar,TWC16_DistributedStochastic_ZandiDongGrami}. The same receiver can also detect other transmissions by keeping her transmitter silent. Motivated by CRN applications, we refer to this feature as`sensing'. When a player senses another player on the selected arm, she refrains form playing it and receives zero reward, but locked player receives reward as no collision occurs. Also, in the dynamic case, any arm released by leaving players should be taken over by others locked on lower ranked arms. We incorporate these aspects in the ST algorithm to account for regret due to entering and leaving players and develop a new dynamic variant named as Dynamic Trekking with Sensing (DTS) given in Alg.~(\ref{alg:AlgoSensing}). 

\begin{algorithm}[!h]
	\caption{Dynamic Trekking with Sensing (DTS)} \label{alg:AlgoSensing}
	\begin{algorithmic}
		\STATE	Input: {$K, \delta,T, x$ }
		\STATE Compute $\tilde{T}_0$ as in (\ref{eqn:DTS_T0}) and $\tilde{T}_l$ as in (\ref{eqn:DTS_TL})
		\STATE	$(\hat{\mu}, I)$= LearningS $(\tilde{T}_0, K)$
		\STATE	 CTrekkingS$( \hat{\mu}, I)$ 
	\end{algorithmic}
\end{algorithm}

\begin{algorithm}[!b]
	\label{alg:TrekS}
	\caption*{\textbf{Subroutine:} CTrekking}
	\begin{algorithmic}[1]
		\STATE Input : \hspace{-.1cm}{$\pi_1$ (ordered list of estimated arms), $J$ (arm index)}, $\pi_2=[K] \backslash \pi_1$ 
		\STATE Set $C=0, S= 0, L=0, R=1, J_n=1, J_r=J, e=~0$  
		\FOR {$t=T_{0}+1 \dots $}
		\IF{locked ($L==1) $}
		\IF {$S\leq T_{l}$}
		\STATE Stay \& update count $I_t\leftarrow I_{t-1},S \leftarrow S+1$
		\ELSE
		\IF {$I_{t-1}$ is in $\pi_2$}
		\STATE Estimate mean of arm $I_{t-1}$
		\STATE Move $I_{t-1}$ from $\pi_2$ to $\pi_1$ and reorder $\pi_1$
		\ENDIF
		\STATE Return and unlock $R\leftarrow 1, L\leftarrow 0, I_t \leftarrow I_{t-1}, e\leftarrow 0$ 
		\STATE Update reserved arm $J_r \leftarrow I_{t-1}$, set  $J_n \leftarrow 1$
		\ENDIF
		\ELSIF { returned to reserved arm ($R==1$)}
		\IF {$\pi_2$ is empty or $|\pi_2|=e$}
		\STATE Select the next best arm $I_t \leftarrow \pi_1(J_n)$ and sense, 
		\STATE Update next best arm $J_n \leftarrow J_n +1$
		\ELSE
		\STATE Set $e\leftarrow e+1$, select $e$th arm in $\pi_2,I_t \leftarrow \pi_2(e)$
		\ENDIF
		\STATE Set $R \leftarrow 0$
		\IF { all better arms checked ($J_n \geq J_r$)}
		\STATE	Set $L\leftarrow 1, S \leftarrow 0, I_t\leftarrow J_r$,  jump to line (3)
		\ELSIF {locked player is sensed}
		\STATE return $R \leftarrow 1$  
		\ENDIF
		\ELSIF{collision ($\eta_{I_{t-1},{t-1}}==1$)}
		\IF {back-off time lapsed ($C\leq K-J_r$)}
		\STATE stay on the same arm $I_t \leftarrow I_{t-1}$
		\ELSE
		\STATE return to reserved arm $I_t\leftarrow J_r, R\leftarrow 1$ 
		\ENDIF
		\ELSE 
		\STATE  set $L=1$  (lock on $J_n$)
		\ENDIF
		\IF{ collision ($\eta_{I_t,t}==1$)}
		\IF {$L==1$}
		\STATE return to reserved arm $I_t \leftarrow J_r, S \leftarrow 0$
		\ELSE
		\STATE Increase collision count $C \leftarrow C+1$
		\ENDIF
		\ENDIF
		\ENDFOR
	\end{algorithmic}
\end{algorithm}	

DTS also runs in two phases, namely learning phase (Learning) and continuous trekking phase. The learning phase is the same as in ST except that the players sense the selected arm and play it only if no other player is detected. After the learning phase, each player estimates mean of arms for which at least $T_l$ (specified later)  observations are available and estimates of other arms is set to zero. 

The continuous trekking phase of DTS is based on TrekD subroutine discussed in Section~\ref{sec:mtrek}. It allows players to take up any good arms freed up by leaving players earlier. If many players select the same arm simultaneously,  its inbuilt back-off mechanism resolves who will lock on the arm.  Its pseudo-code is given in Subroutine CTrekking. A player in this phase can be in two states namely, locked or trekking-- in the locked state, the same arm is played for $T_l$ rounds. In the trekking state, availability of better arms is checked. The states alternate for each player. 

When a player enters the CTrekking phase from learning phase, she may not have estimates of all the arms. Let $\pi_1$ denote the set of arms for which a players has good estimate of mean rewards. We first explain CTrekking for the case $|\pi_1|=K$, i.e., estimates for all the arms is available and then explain how to account for the other case where estimates are unknown, i.e., $|\pi_1|<K$.  When $|\pi_1|=K$, CTrekking is exactly same as the TrekD, except that a player enters into trekking state if she in locked state for $T_l$ rounds on an arm and returns to the reserved arm if she observes a collision while in locked state.  When $|\pi_1|<K$, the TrekD subroutine has to be modified so that the players estimate the arms in  $\pi_2:=[K] \backslash \pi_1$ .


The case $|\pi_1|<K$ can happen for a new player who cannot get to observe some arms as they could be occupied by the other players already in the game. Since the occupied arms are likely to be the top arms, during the CTrekking phase the new player should check their availability. This allows the player to take-over one of the top arm as soon as they are freed-up. Specifically, after entering into the CTrekking phase, she sets arm $K$ as her reserved arm. In this case the procedure of sensing and locking on an arm is same as the case with $|\pi_1|=K$ with the following differences in the way an arm is selected. Each time the player enters into trekking state she first checks the arms for which the estimates are not available. If she locks on any of these arms and plays it for $T_l$ rounds, she estimates its mean and moves the arm from the set of of un-estimated arms ($\pi_2$) to the set of estimated arms ($\pi_1$). Once all the un-estimated arms are checked it selects the arms from the estimated arm according to their rank.


\begin{thm}
	\label{thm:DynamicDTS}
	Consider the same setup as in Prop. \ref{prop:DynamicDT}. For any $\delta\in (0,1)$, setting duration of learning phase ($\tilde{T}_0$) and  locking period $(T_l)$ as 
	\begin{equation}
	\label{eqn:DTS_T0}
\hspace{-.3cm}		\tilde{T}_0=\left \lceil \frac{\log(\delta/2(K+x))}{\log\left(1-1/4K\right)}\right \rceil + \frac{2K}{\epsilon^2}\log \left (\frac{4K(K+x)}{\delta}\right).
	\end{equation}
	\begin{equation}
	\label{eqn:DTS_TL}
	T_l=\sqrt{T\tilde{T}_{tr}/x}.
	\end{equation}
the regret and collision in DTS are bounded with probability at least $1-\delta$ as
	$R\leq \mathcal{O}(\sqrt{xT}) \mbox{  and  } C\leq \mathcal{O}(\sqrt{xT}).$
\end{thm}

In DTS, constant regret is incurred during the learning phase and it grows with  time in the continuous trekking phase. Since the players have to periodically check availability of better arms, the regret increases with the duration of stay. The proof of Thm $4$ first bounds the regret due to all types of players during the learning and trekking phase and then optimizes over the rate at which the players should check for availability of better arms. The detailed proof is given in the appendix. Note that the length of the learning phase $\tilde{T}_0$ is a constant and does not grow with $T$.    

The DTS algorithm does not require a global clock as in the case of DT and DMC hence is completely decentralized. Further, players can enter and leave at any time during the game  which eliminates any need to regulate the operations of the players. However, DTS still requires to know the horizon $T$ to achieve sub-linear regret. Relaxing this requirements is still an open challenge for future work.

\section{Simulation Results}
\label{sec:SR}
\begin{figure*}[t]
	\centering     
	\subfigure[]{\label{st11}\includegraphics[width=50mm]{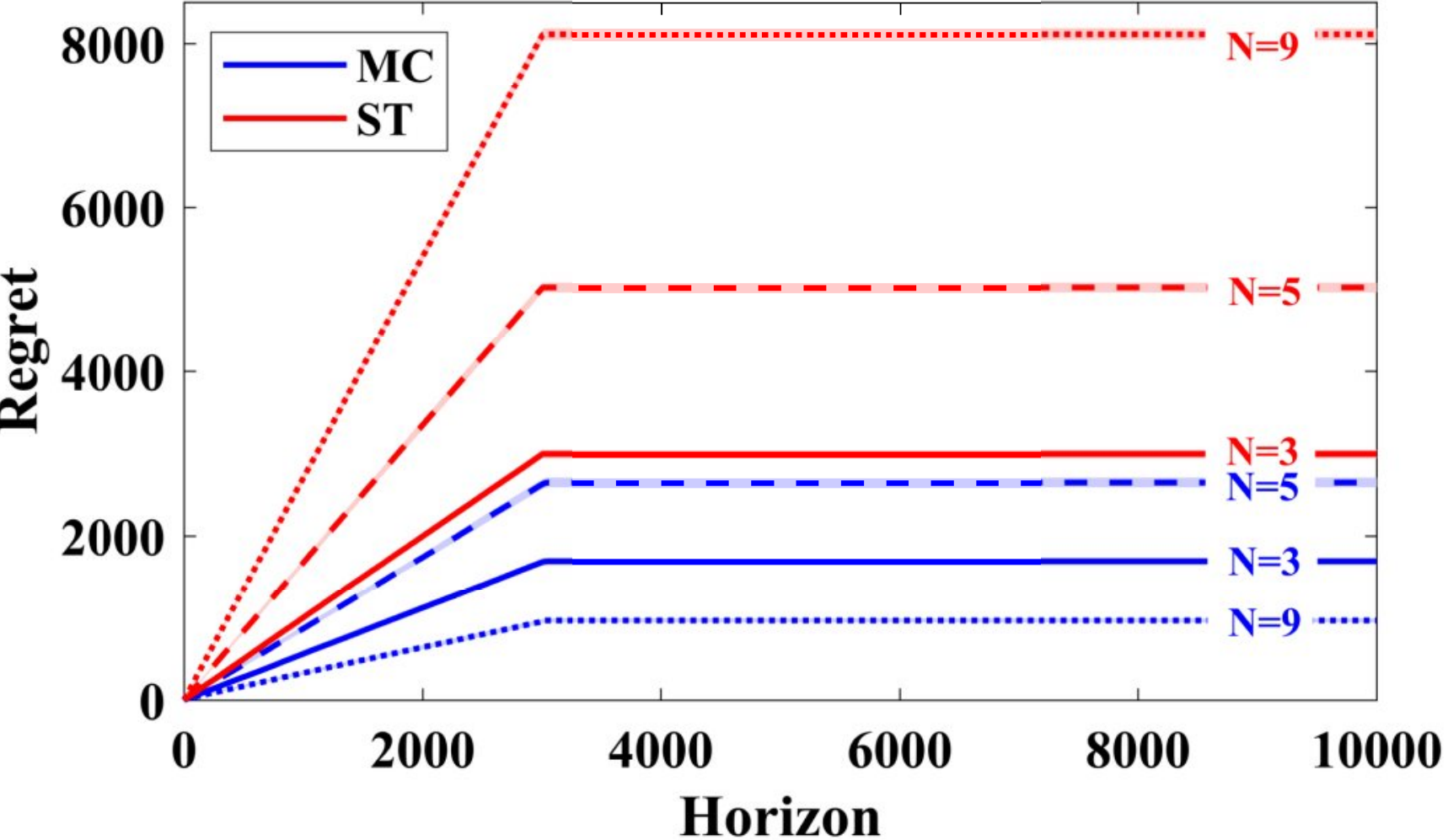}} \hspace{0.25cm}
	\subfigure[]{\label{st12}\includegraphics[width=55mm]{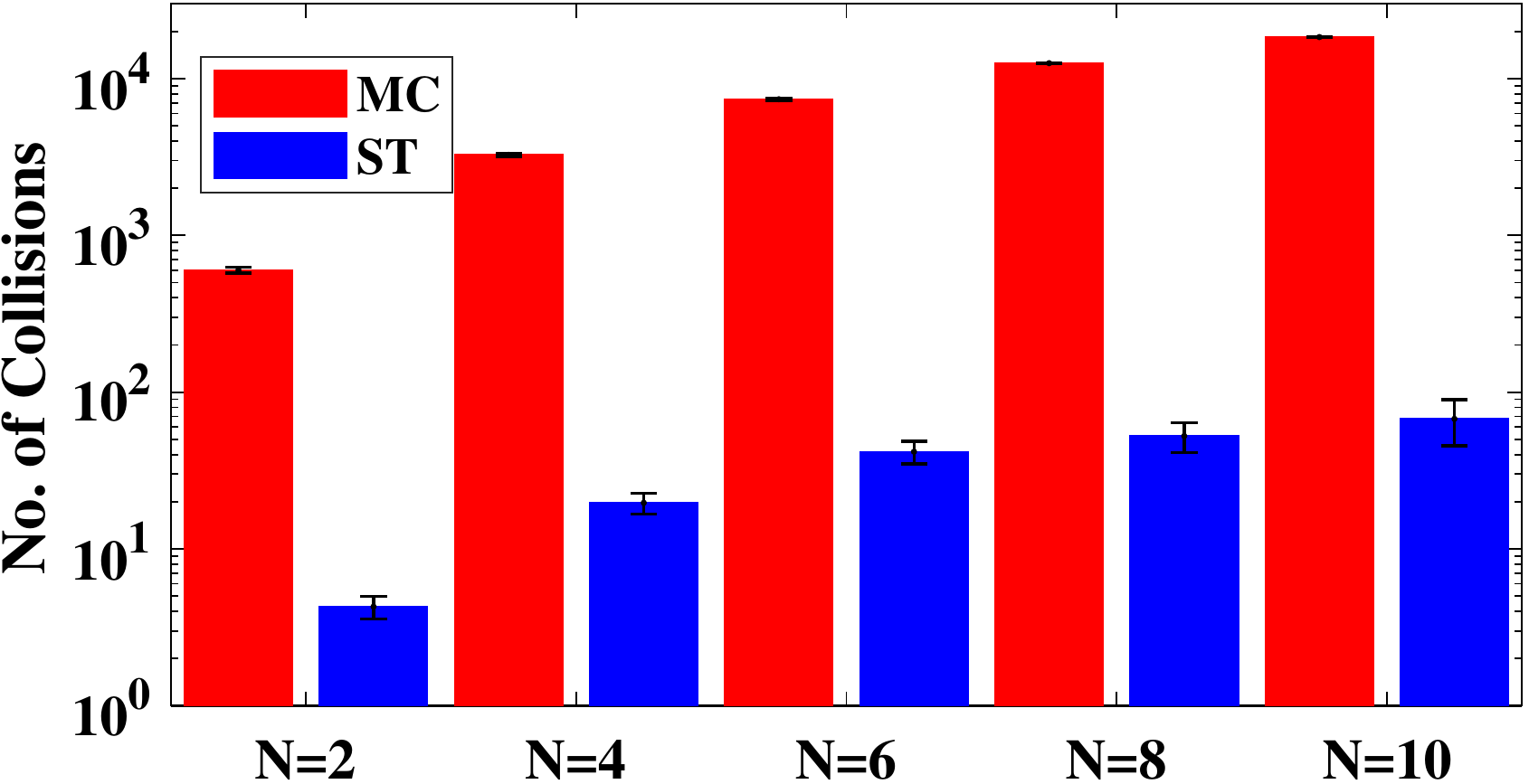}}\hspace{0.25cm}
	\subfigure[]{\label{st13}\includegraphics[width=50mm]{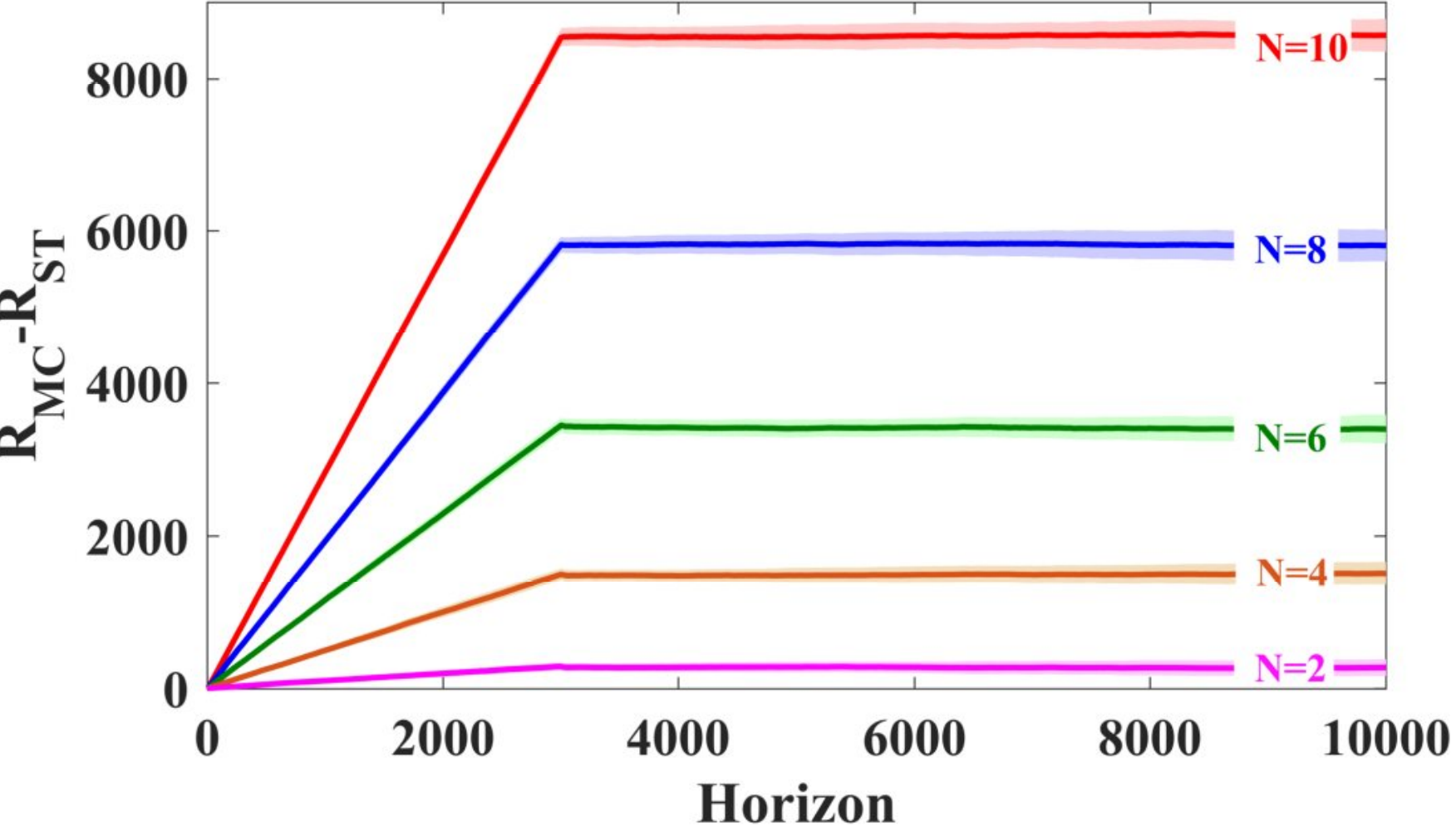}}
%
	\caption{(a) The cumulative regret comparison between MC and ST algorithm for $N=4$ and $N=8$, (b) Number of collisions for different values of $N$ with $y-axis$ shown on the logarithmic scale, and (c) The plots showing the difference between the regret of MC and ST algorithms for various values of $N$. Here, we consider the arms with statistics $\mu^1$, $T_0=T_0^{MC}=3000$.}
	
	\label{st1}
	
\end{figure*}
\begin{figure*}[!t]
	\centering     
	\subfigure[]{\label{st21}\includegraphics[width=55mm]{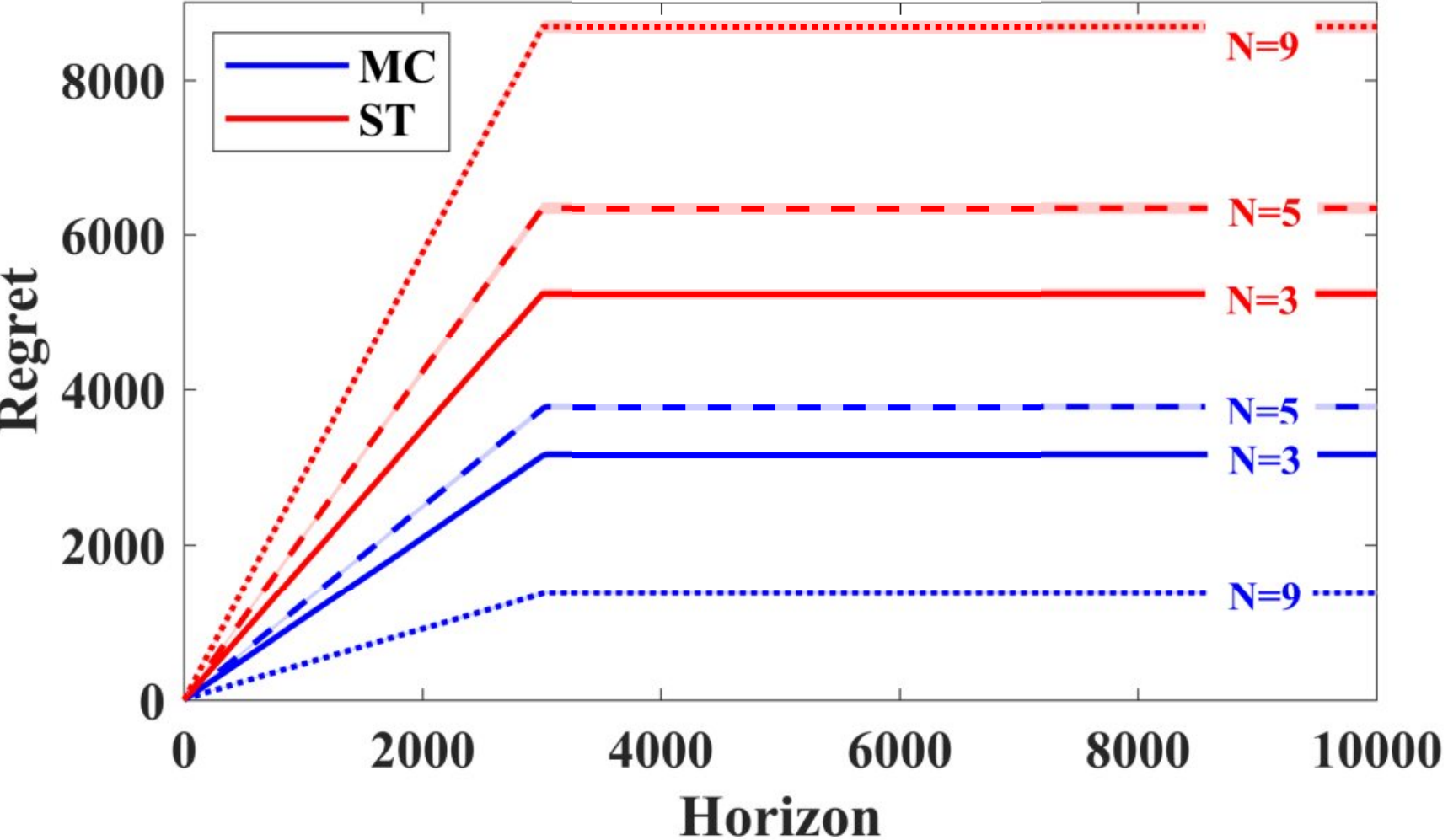}} 
	\subfigure[]{\label{st22}\includegraphics[width=58mm]{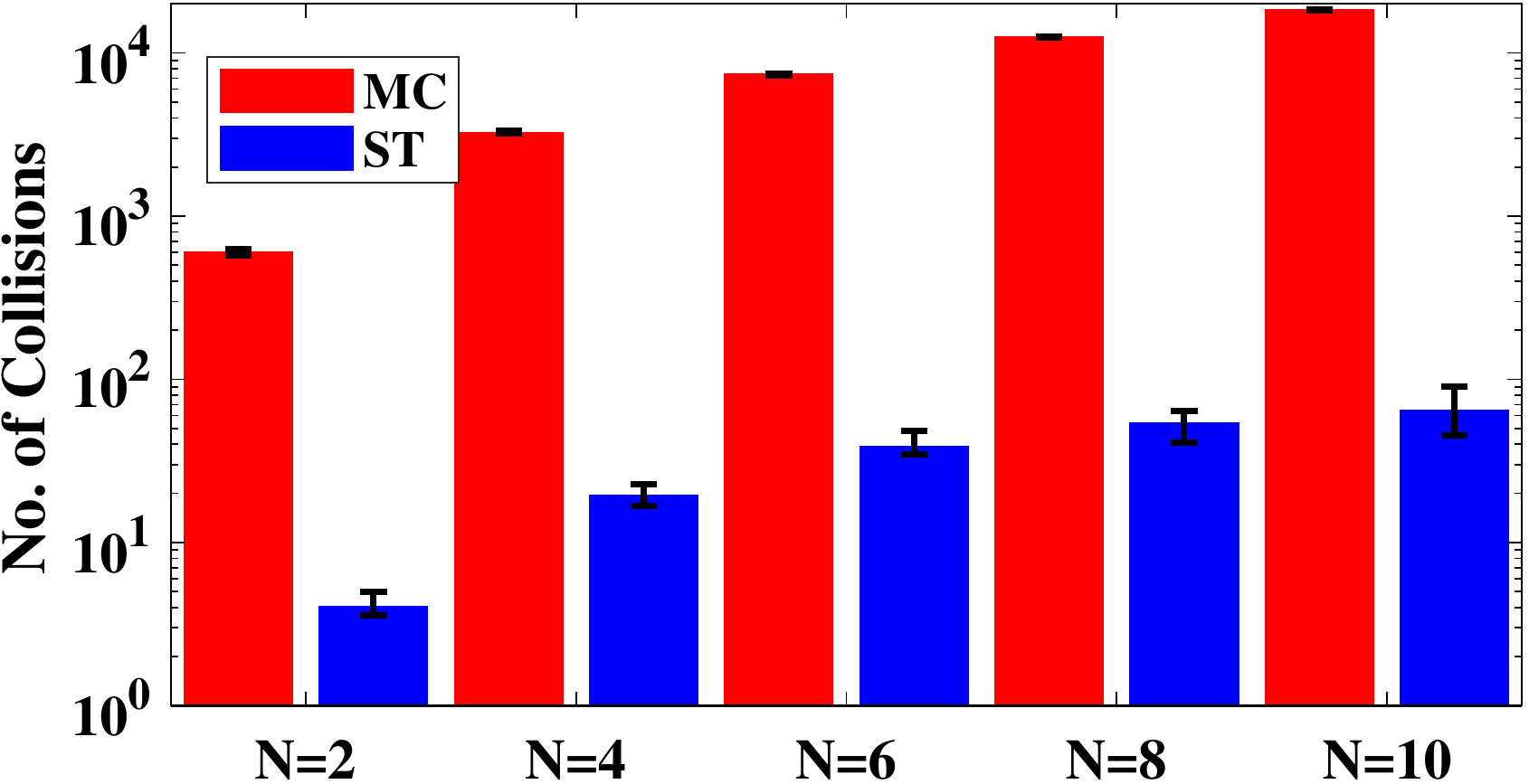}}\hspace{0.1cm}
	\subfigure[]{\label{st23}\includegraphics[width=55mm]{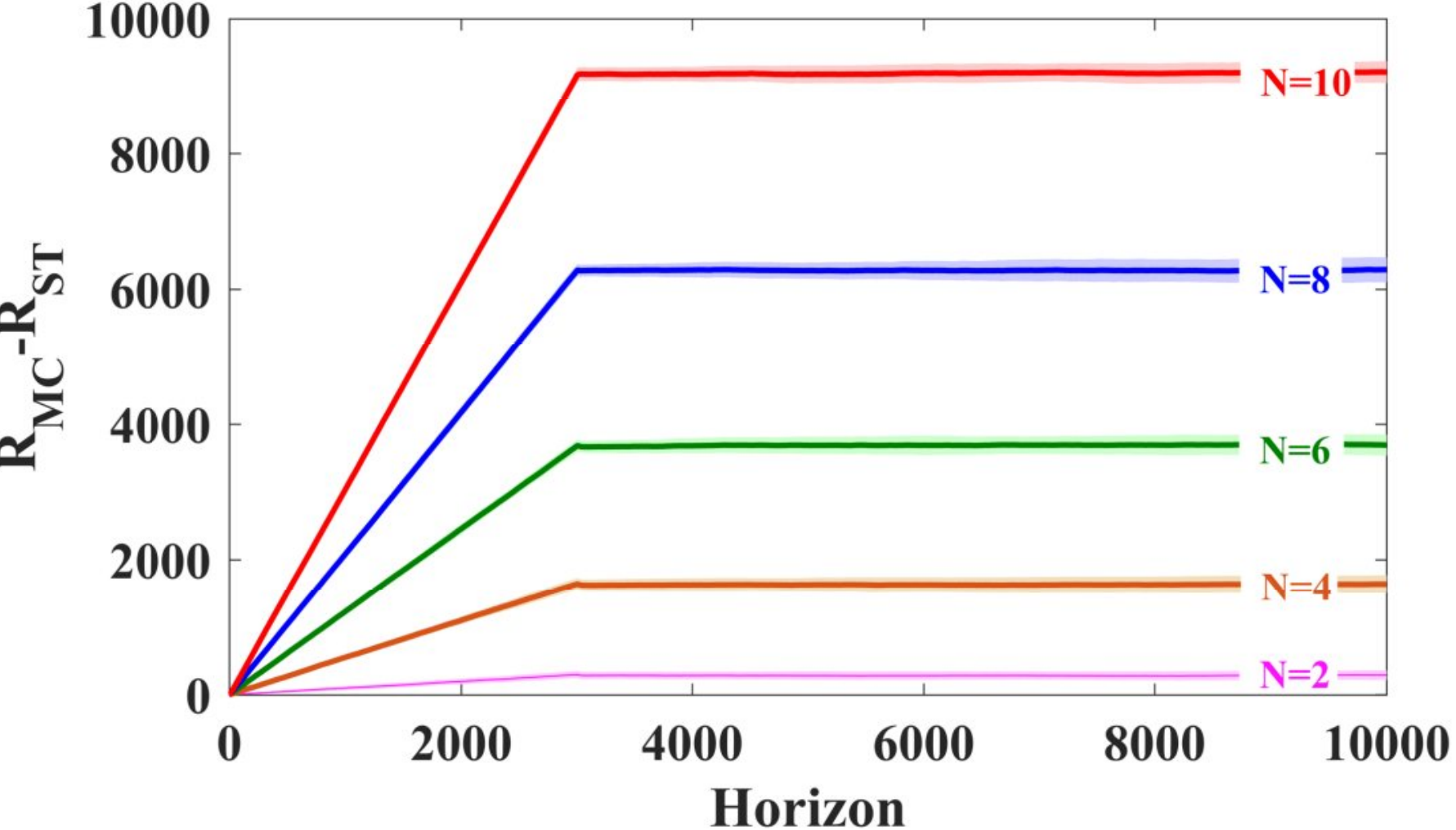}}
		\vspace{-0.2cm}
	\caption{(a) The cumulative regret comparison between MC and ST algorithm for $N=4$ and $N=8$, (b) Number of collisions for different values of $N$ with $y-axis$ shown on the logarithmic scale, and (c) The plots showing the difference between the regret of MC and ST algorithms for various values of $N$. Here, we consider the arms with statistics $\mu^2$, $T_0=T_0^{MC}=3000$.}
	\label{st2}
\end{figure*}

We implemented the ST algorithm for the static case and the DT and DTS algorithms for dynamic case. For comparison, we implemented MC and DMC algorithms in \cite{ICML16_MultiplayerBandits_RosenkiShamir}, which are the current state-of-the-art for the static and dynamic cases, respectively. These algorithms have shown to outperform algorithms in \cite{JSAC11_DistributedLearning_Anadakumar,TSP10_DistributedLearning_LiuZhao,TSP14_DistributedStochastic_GaiKrishnamachari,TWC16_DistributedStochastic_ZandiDongGrami,KDD14_ConcurrentBandits_AvnerMannor} and hence, we do not include them here for better clarity of the plots. The parameters of the MC and DMC algorithms are chosen as suggested in \cite{ICML16_MultiplayerBandits_RosenkiShamir}, to achieve best possible regret. 

Mean rewards of $N$ arms are set such that $\mu_{\lceil \frac{N}{2}\rceil} = 0.5$ and for $n > {\frac{N}{2}}$ and $n < {\frac{N}{2}}$, the gap between the means of $n^{th}$ and $(n+1)^{th}$ arms is at least 0.05 as suggested in \cite{ICML16_MultiplayerBandits_RosenkiShamir}. We consider the various set of means depicting various scenarios in static and dynamic cases. For each setup and algorithm, the experiments are repeated $50$ times and each plot includes the cumulative regret, average regret and standard deviation (shown with a shaded region). In the dynamic case, we mark the rounds at which player enters or leaves with orange dashed and gray dash-dot lines, respectively. We also compare the average number of collisions faced by all the players during the game.


%
%

%
%
%

%

\subsection{Static Case}
For static case, we consider a game of $T=10000$ rounds. We consider $K=10$ and $\mu^1 =  \{0.22,0.29,0.36,0.43,0.50,0.57,0.64,0.71,0.78,0.85\}$ with $\Delta=0.07$ and  $\mu^2 =  \{0.05,0.15,0.25,0.35,0.45,0.55,0.65,0.75,0.85,0.95\}$ with $\Delta=0.1$. Analytically, the $T_0$ of the ST algorithm is at least $4$ times smaller than $T_0^{MC}$ of the MC algorithm. However, $T_0^{MC}$ used in the simulation results presented in \cite{ICML16_MultiplayerBandits_RosenkiShamir} is much smaller than that obtained from their mathematical expressions. For fair comparison, we assume $T_0^{MC}=T_0=3000$ rounds. Later, we present the results using actual values of $T_0^{MC}$ and $T_0$.

%
%

In Fig.~\ref{st11}, we compare the cumulative regret of the MC and ST algorithms for $N=\{3,5,9\}$ where $N$ indicates the number of active players. The MC algorithm has constant regret plot after learning and MC phases while the ST algorithm has constant regret plot after learning and trekking phases. The constant regret plot is an indication of players settling in top channels and no further increase in the regret thereafter. Before that, the regret of MC algorithm is significantly higher than that of the ST algorithm due to random hopping approach in the MC algorithm compared to collision-free sequential hopping in the ST algorithm. Also, the regret of the ST algorithm is highest when $N=K/2=5$ while the regret of the MC algorithm increases as the value of $N$ increases. 

Next, we compare the number of collisions faced by all the players until the end of the horizon in Fig.~\ref{st12}. Note that $y-axis$ is shown on a logarithmic scale for clarity of the plots. It can be observed that the number of collisions is significantly higher in the MC algorithm compared to the ST algorithm and the difference increases as the value of $N$ increases. This, in turn, means that the difference between the regret of the MC algorithm and ST algorithm increases as the value of $N$ increases. The corresponding plots are shown in Fig.~\ref{st13}. Similarly, the plots corresponding to $\mu^2$ are shown in Fig.~\ref{st2}. All the plots presented in Fig.~\ref{st1} and Fig.~\ref{st2} validate our claims in Section~\ref{staticthm}.

\begin{figure*}[!t]
	\centering     
	\subfigure[]{\label{st31}\includegraphics[width=55mm]{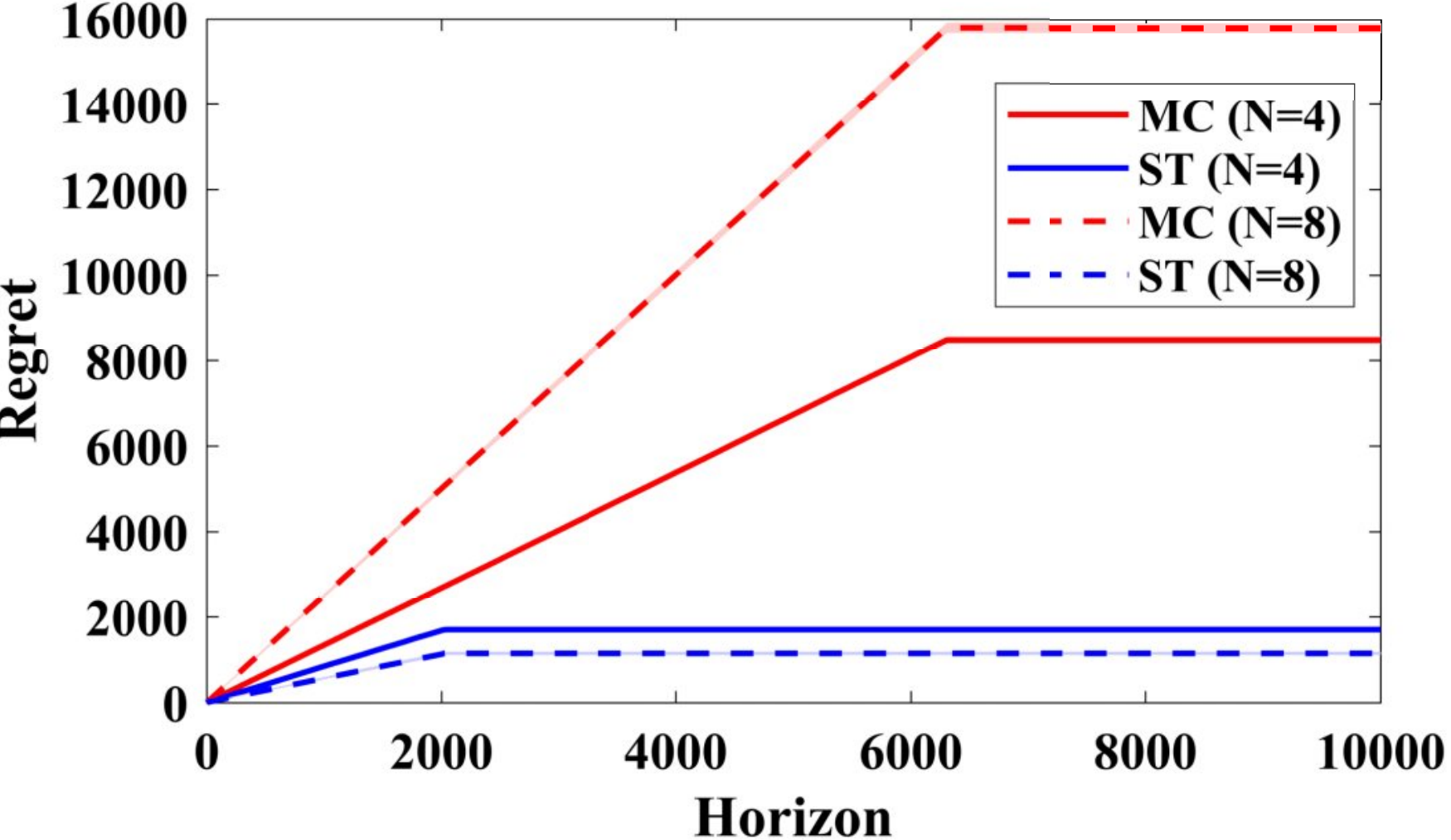}}
	\subfigure[]{\label{st32}\includegraphics[width=58mm]{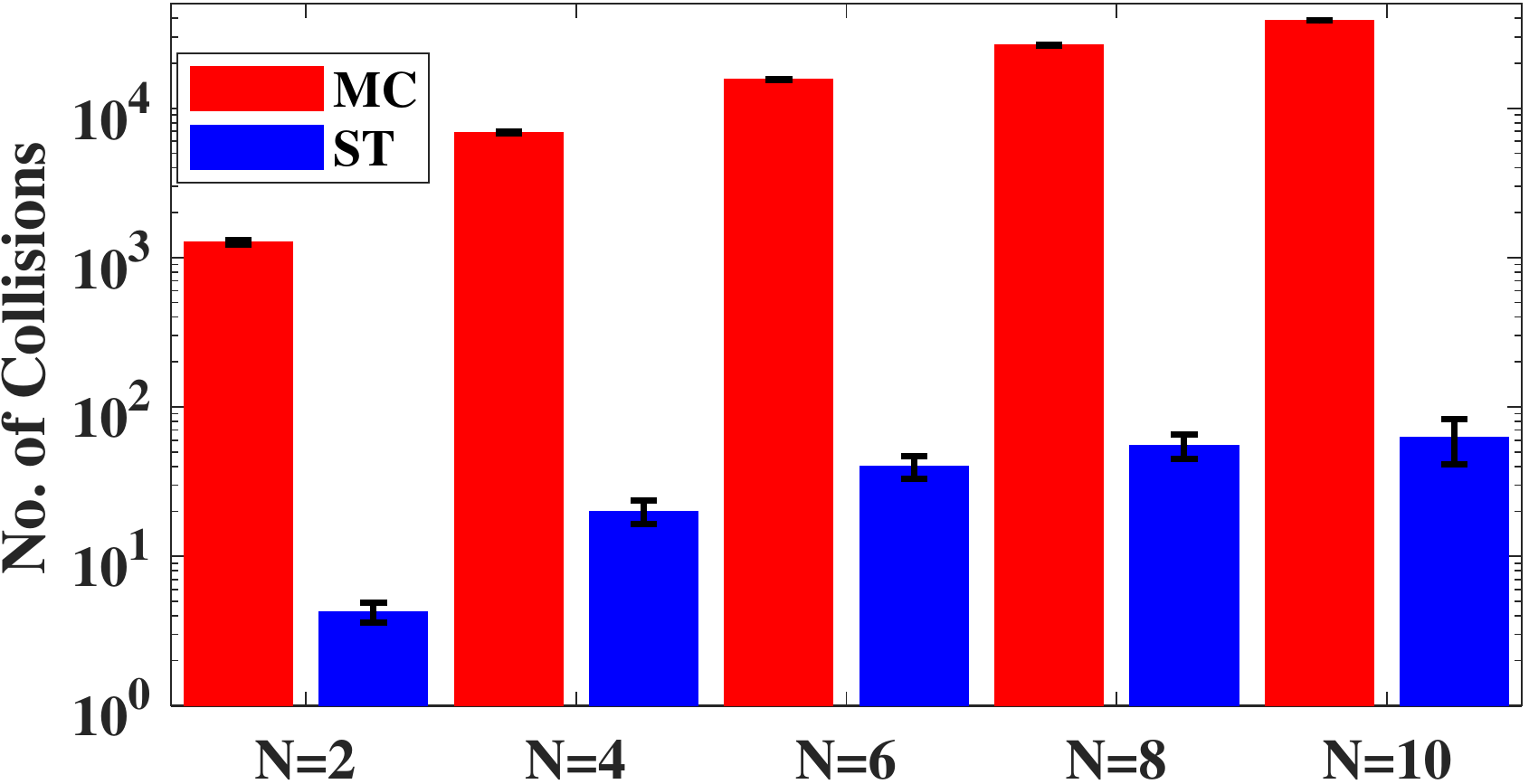}}\hspace{0.1cm}
	\subfigure[]{\label{st33}\includegraphics[width=55mm]{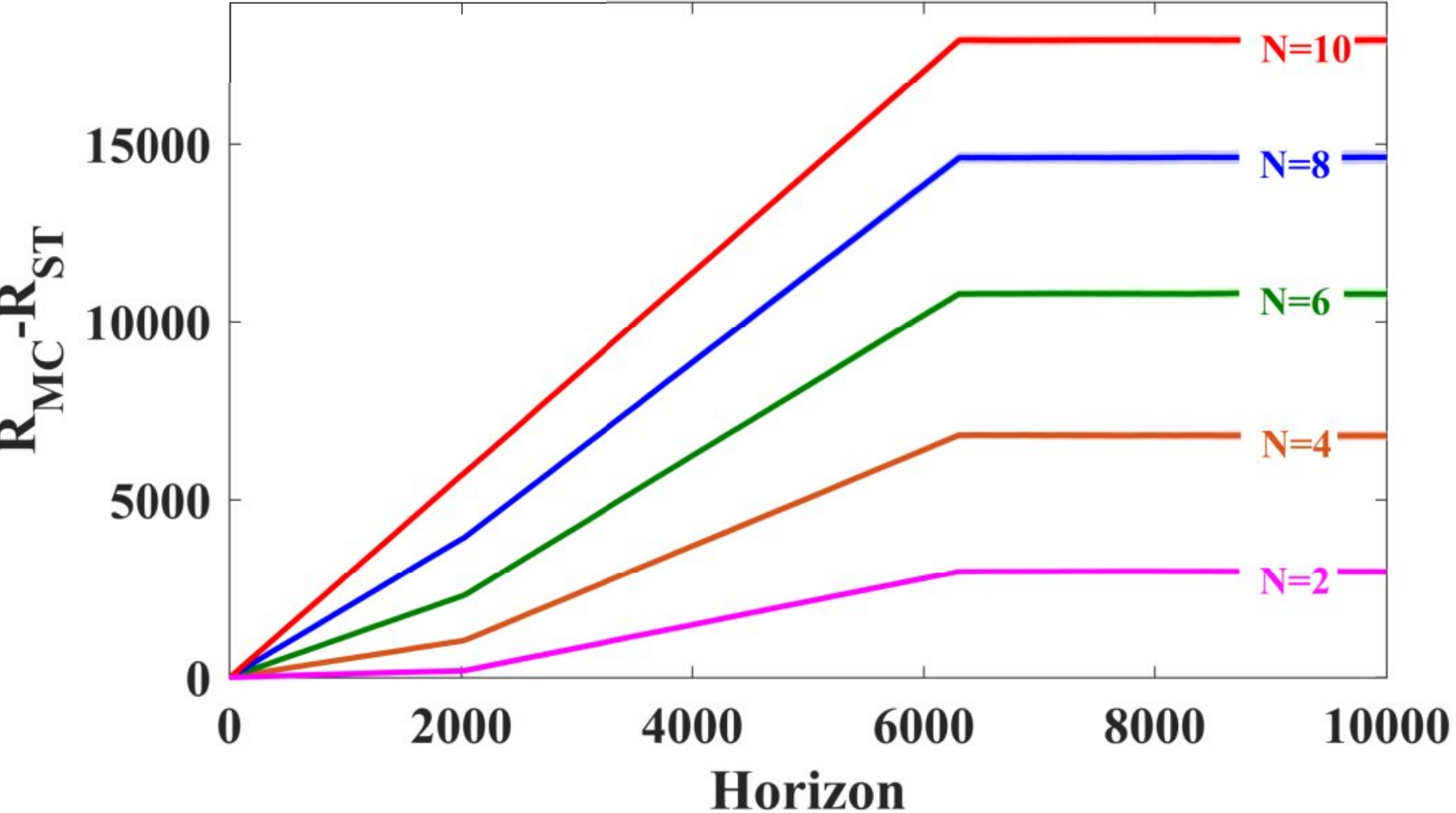}}
		\vspace{-0.2cm}
	\caption{(a) The cumulative regret comparison between MC and ST algorithm for $N=4$ and $N=8$, (b) Number of collisions for different values of $N$ with $y-axis$ shown on the logarithmic scale, and (c) The plots showing the difference between the regret of MC and ST algorithms for various values of $N$. Here, we consider the arms with statistics $\mu^1$, $T_0=2000$ and $T_0^{MC}=6200$.}
	\label{st3}
\end{figure*}

\begin{figure*}[!t]
	\centering     
	\subfigure[]{\label{st41}\includegraphics[width=55mm]{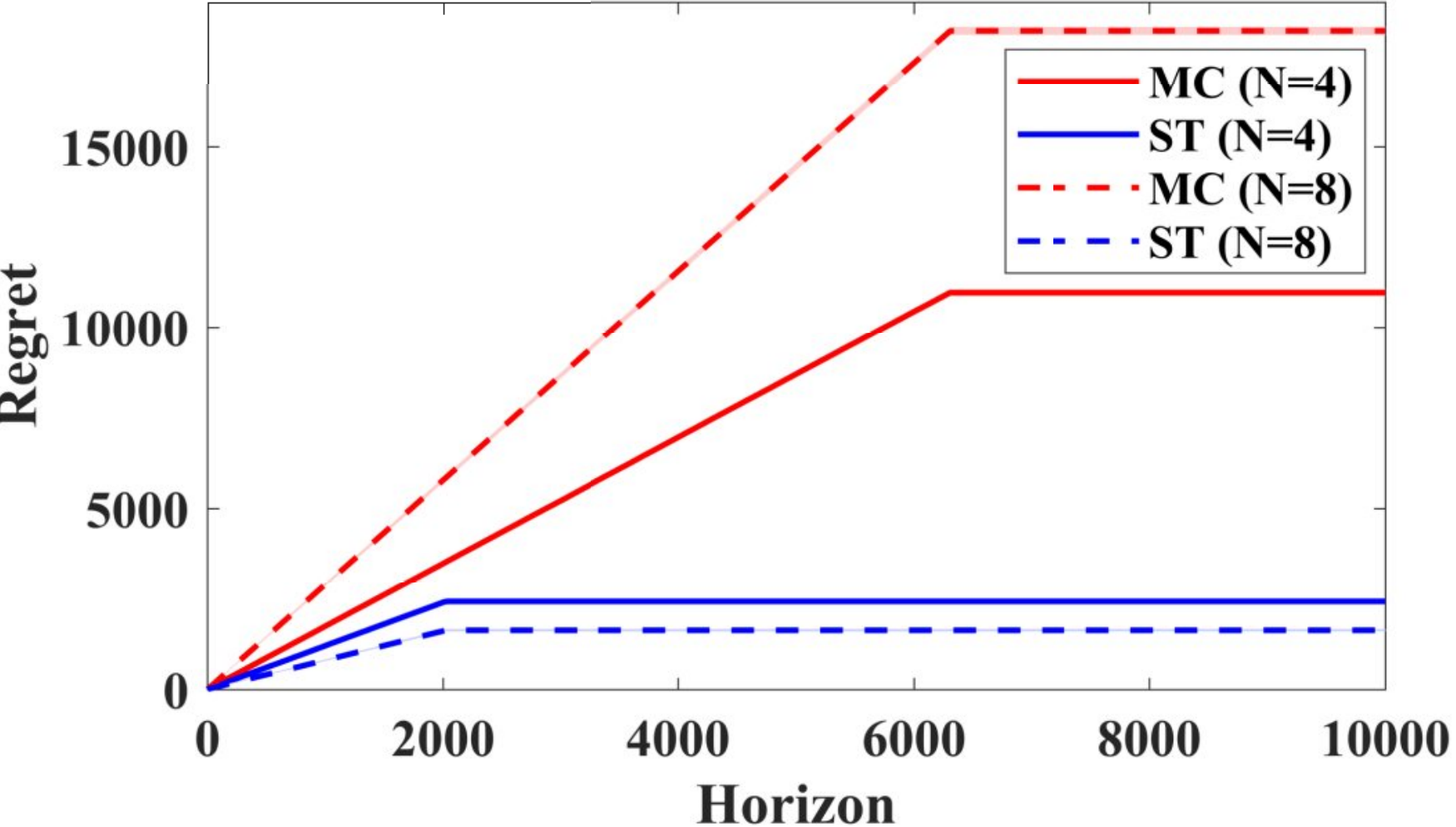}} 
	\subfigure[]{\label{st42}\includegraphics[width=58mm]{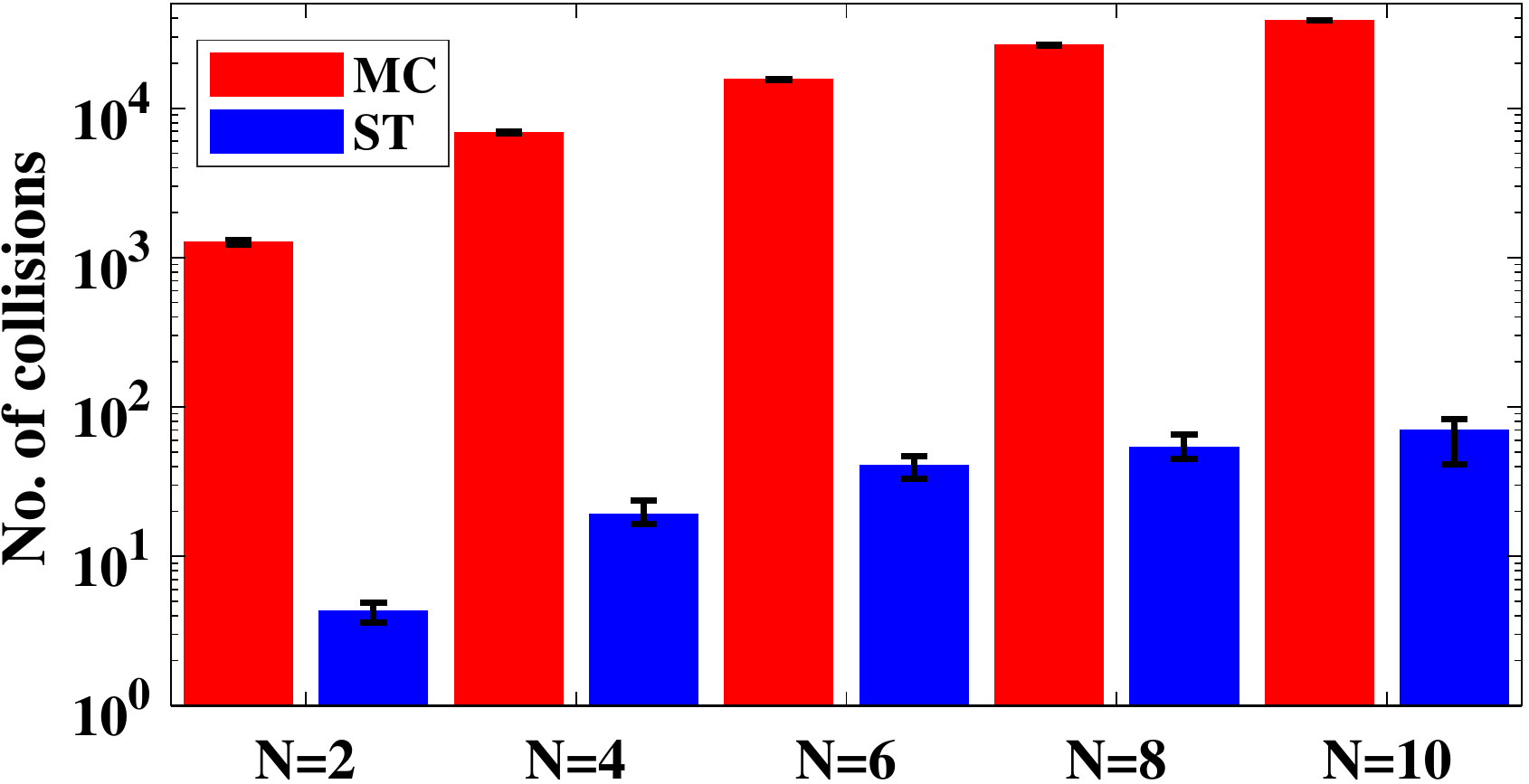}}\hspace{0.1cm}
	\subfigure[]{\label{st43}\includegraphics[width=55mm]{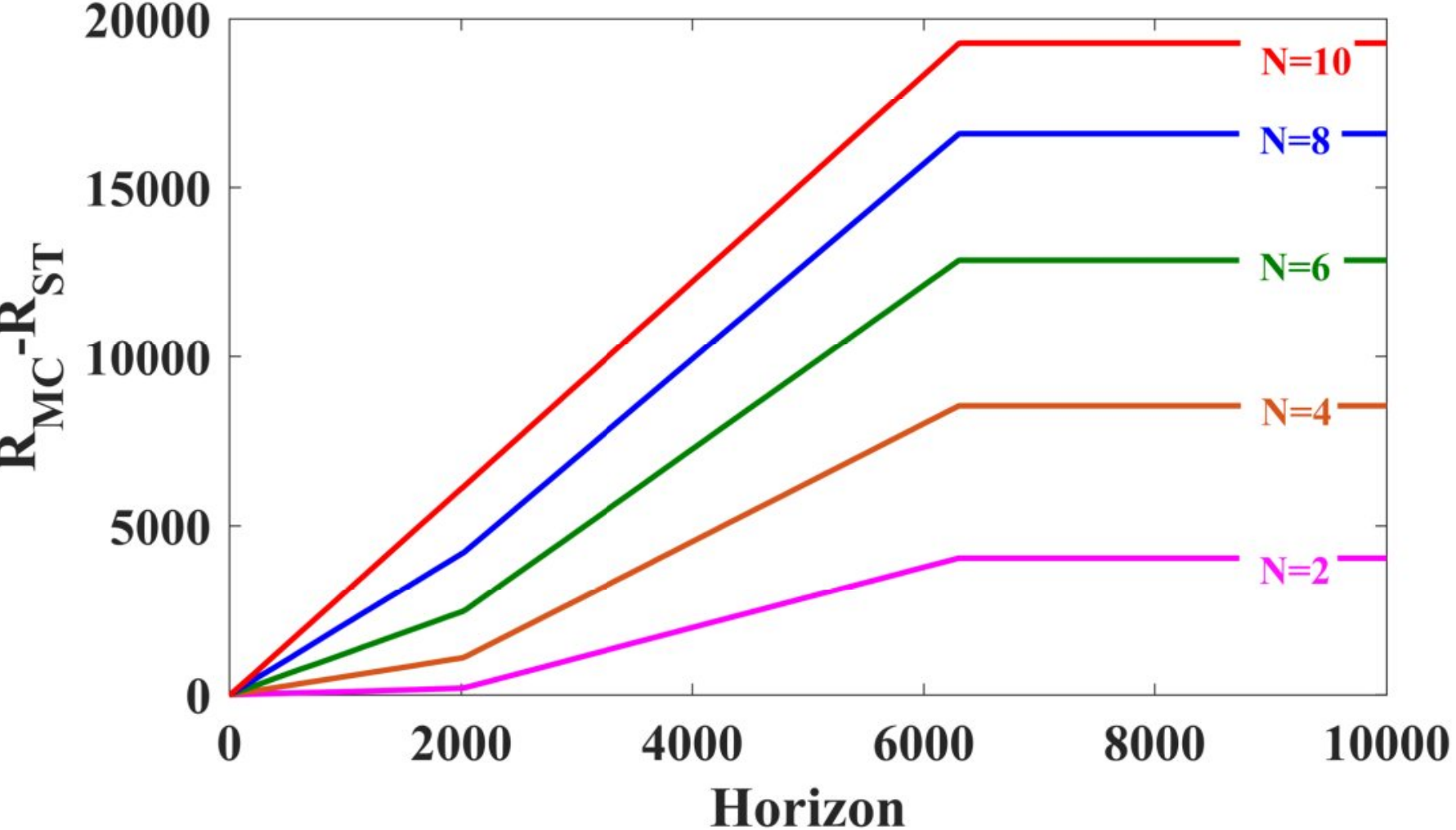}}
	\vspace{-0.2cm}
	\caption{(a) The cumulative regret comparison between MC and ST algorithm for $N=4$ and $N=8$, (c) Number of collisions for different values of $N$ with $y-axis$ shown on the logarithmic scale, and (c) The plots showing the difference between the regret of MC and ST algorithms for various values of $N$. Here, we consider the arms with statistics $\mu^2$, $T_0=2000$ and $T_0^{MC}=6200$.}
	\label{st4}
\end{figure*}

The simulation results presented in Fig.~\ref{st1} and Fig.~\ref{st2} assume $T_0^{MC}=T_0=3000$. As discussed in Section~\ref{sec:AST}, the actual value of $T_0$ for the ST algorithm is much smaller than $T_0^{MC}$ of the MC algorithm. Here, we choose the value of $T_0$ and $T_0^{MC}$ such that it guarantees desired minimum number of observations of each arm, $C_m$ at each player. The value of the $C_m$ is given in Lemma 2 and is equal to 200 for the parameters considered here. The corresponding value of $T_0^{MC}$ and $T_0$ are 6200 and 2000, respectively. The plots of the cumulative regret, and the number of collisions for two different arm statistics, $\mu^1$ and $\mu^2$ are shown in Fig.~\ref{st3} and Fig.~\ref{st4}, respectively. It can be observed that the difference between the regret of ST and MC algorithms is higher in this case compared to the results in Fig.~\ref{st1} and Fig.~\ref{st2} where $T_0^{MC}=T_0$. Similar observations can be made for the number of collisions as well.


\subsection{Dynamic Case}
In this sub-section, we consider various scenarios to compare the performance of the DT, DTS and DMC algorithms. The game starts with one player and the subsequent entry and exit of the players are shown using dotted red and dashed gray lines, respectively. The leaving players are chosen randomly.

\subsubsection{Scenario 1-3: Restricted Entry/Exit}
In this sub-section, we study the effect of the number of players and the rate at which they enter or leave the game. To do this, we consider three scenarios where players can enter or leave the game anytime except during learning period of each epoch of the DMC and DT algorithms. 

We begin with Scenario 1 which is similar to the one discussed in \cite{ICML16_MultiplayerBandits_RosenkiShamir}. In Scenario 1, the game starts with one player and $T=500000$ rounds. At $166667^{th}$ round, second player enters while the first player leaves later at $333333^{th}$ round. Similar to \cite{ICML16_MultiplayerBandits_RosenkiShamir}, we consider four arms with $\mu^3 =\{0.05, 0.35,0.65,0.95\}$ and epoch length, $T_{ep}$ of 34757 rounds. Due to single player till 166667 rounds, the regret of the DMC and DT algorithm is identical while the regret of the DTS algorithm is lower due to epoch-free approach. In fact, the DTS algorithm incurs regret only when the player enters or leaves the game. The regret of the DT algorithm is lower than that of the MC algorithm from 166667 till 333333 rounds during which there are two players in the game. Thereafter, the regret incurred by both the algorithms is identical. Thus, the DT algorithm is superior to the DMS algorithm whenever there are more than one player in the game. 

\begin{figure}[!h]
	\centering     
	\subfigure[]{\label{dc11}\includegraphics[width=40mm]{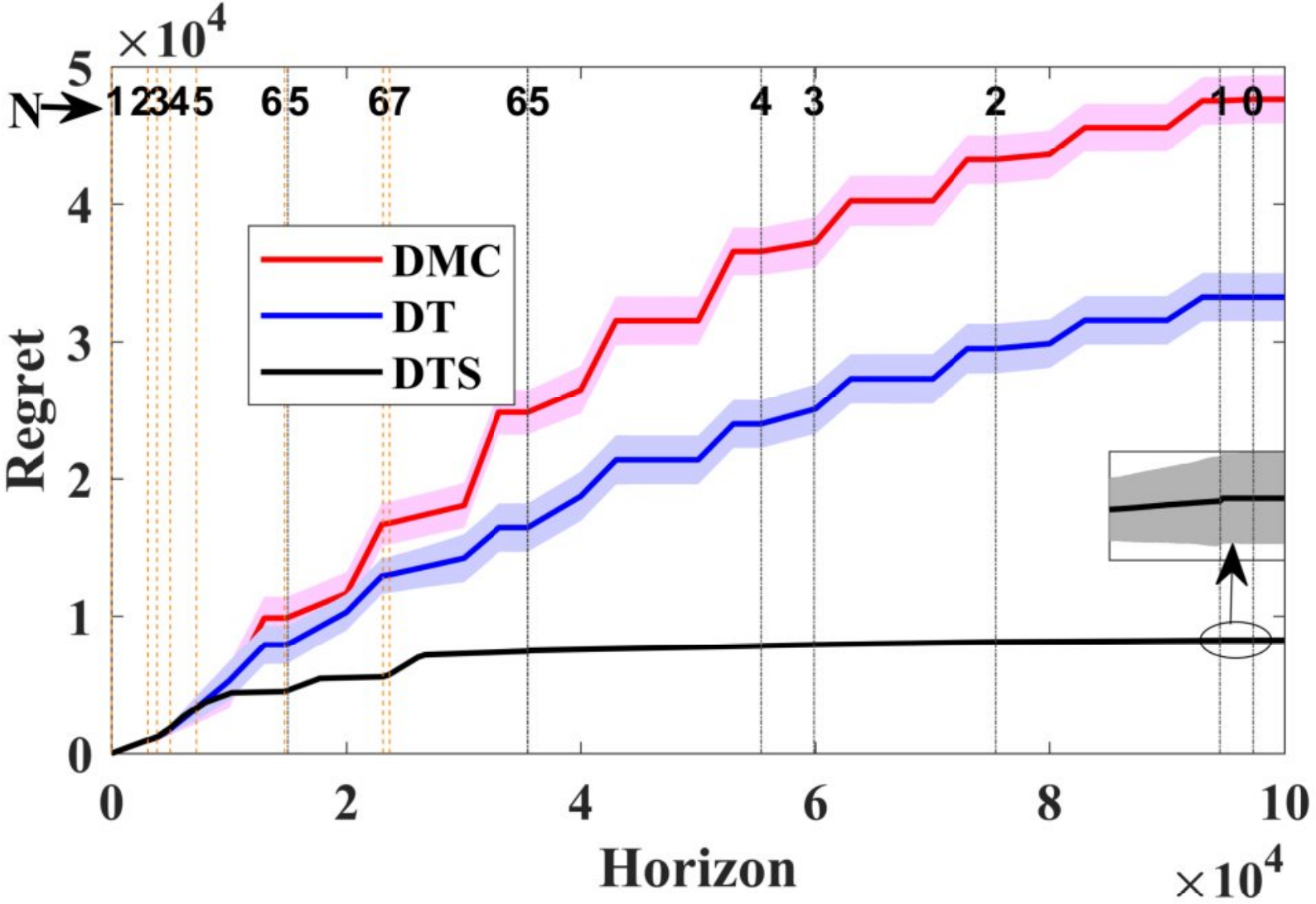}}  \hspace{0.25cm}
	\subfigure[]{\label{dc12}\includegraphics[width=41mm]{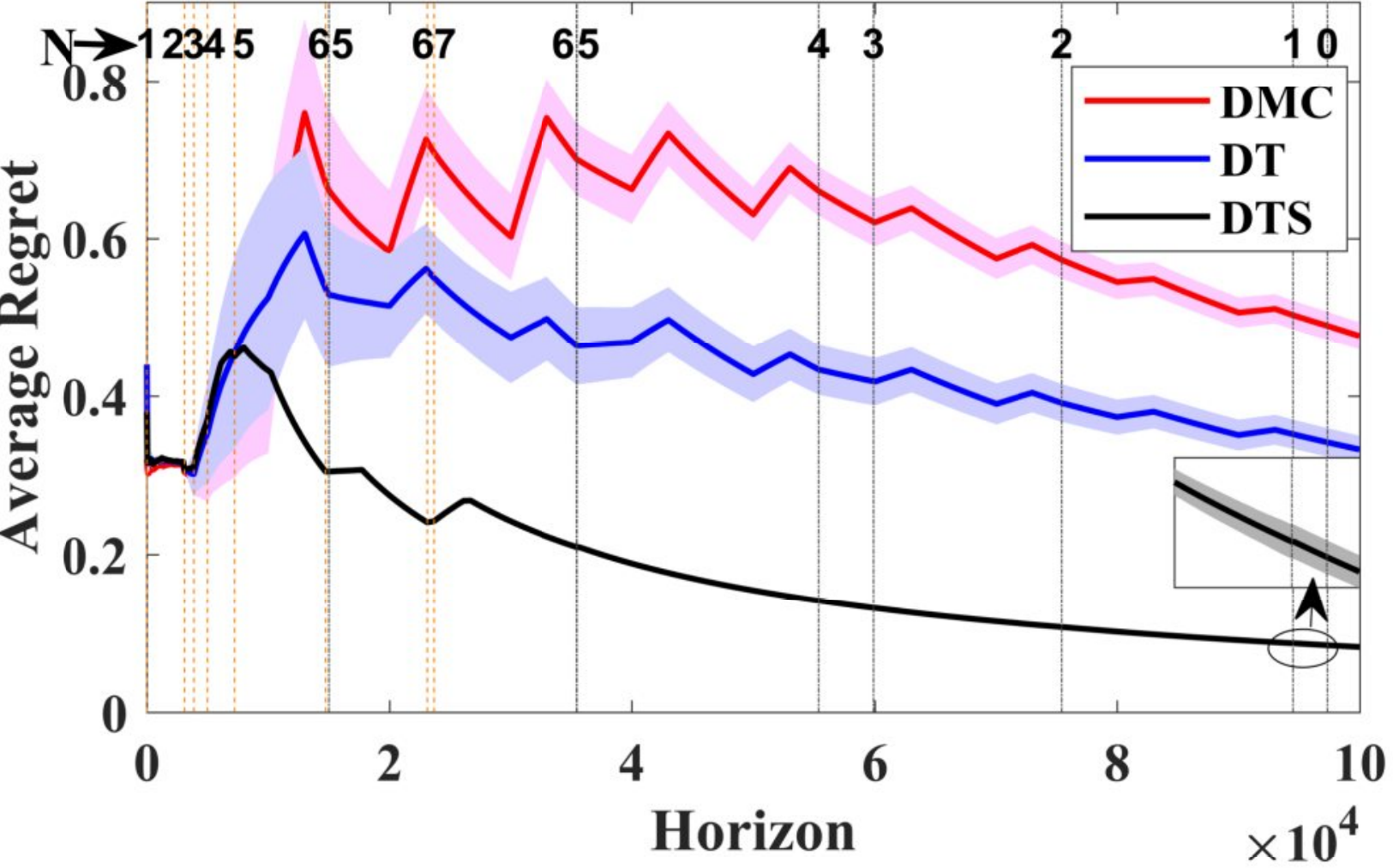}}
	\caption{(a) Cumulative regret and (b) Average regret comparison for Scenario 2 where players can enter or leave any time except during learning phase of the DMC and DT algorithm. }
	\label{dc1}
\end{figure}

\begin{figure}[!h]
	\centering     
	\subfigure[]{\label{dc21}\includegraphics[width=40mm]{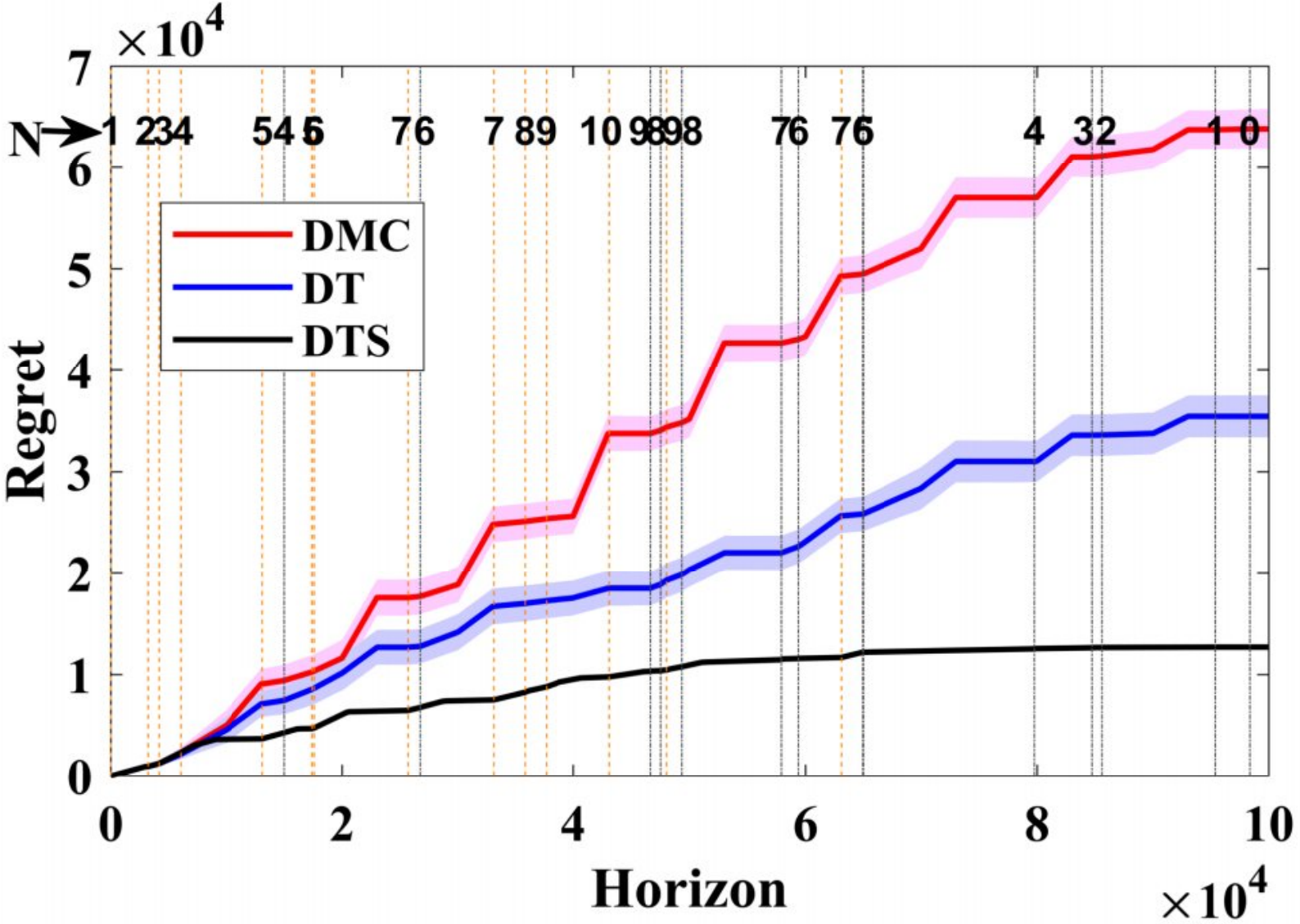}}  \hspace{0.25cm}
	\subfigure[]{\label{dc22}\includegraphics[width=41mm]{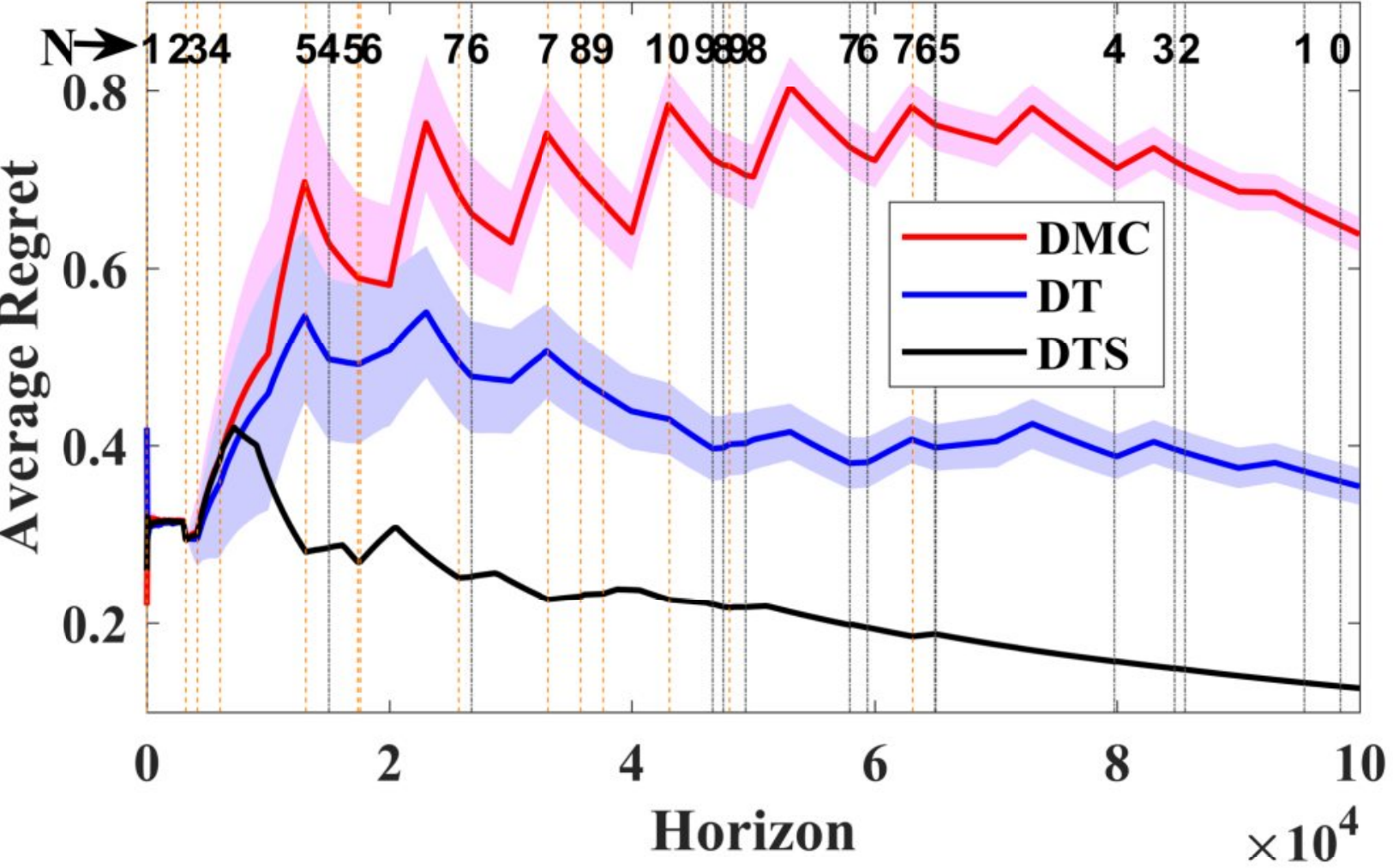}}
	\caption{(a) Cumulative regret and (b) Average regret comparison for Scenario 3 where players can enter or leave any time except during learning phase of DMC and DT algorithm.}
	\label{dc2}
\end{figure}

Next, we increase the number of players and the rate at which they enter or leave the game. The corresponding plots are shown in Fig.~\ref{dc1} and Fig.~\ref{dc2} where the number of players and the entering/leaving rate is higher in latter compared to former. As expected, the regret of the DTS algorithm is lowest followed by that of the DT algorithm and MC algorithm incurs highest regret among three algorithms.

\subsubsection{Scenario 4-5: Un-restricted Entry/Exit}
\begin{figure}[!h]
	\centering     
	\subfigure[]{\label{dc31}\includegraphics[width=40mm]{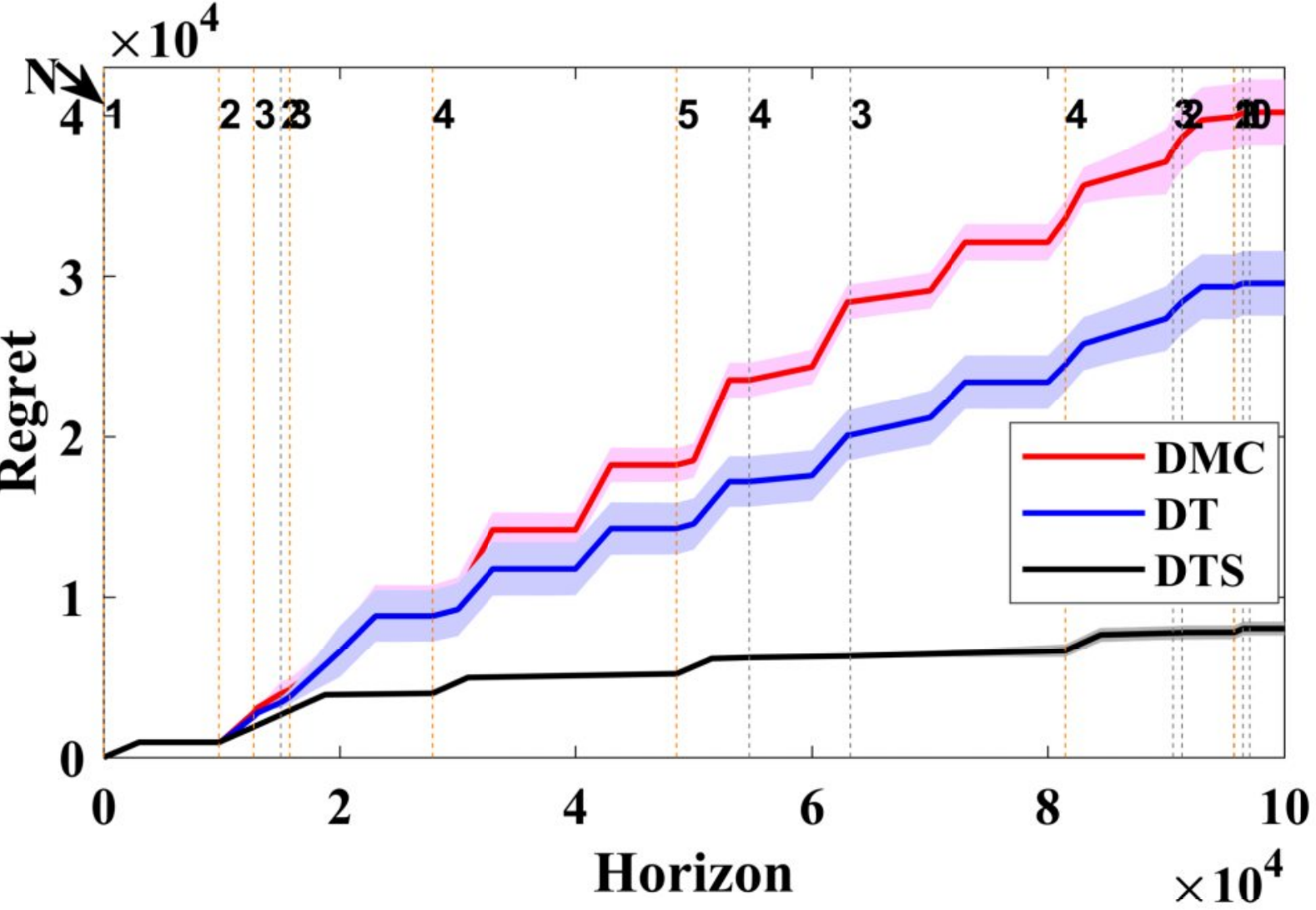}}  \hspace{0.25cm}
	\subfigure[]{\label{dc32}\includegraphics[width=41mm]{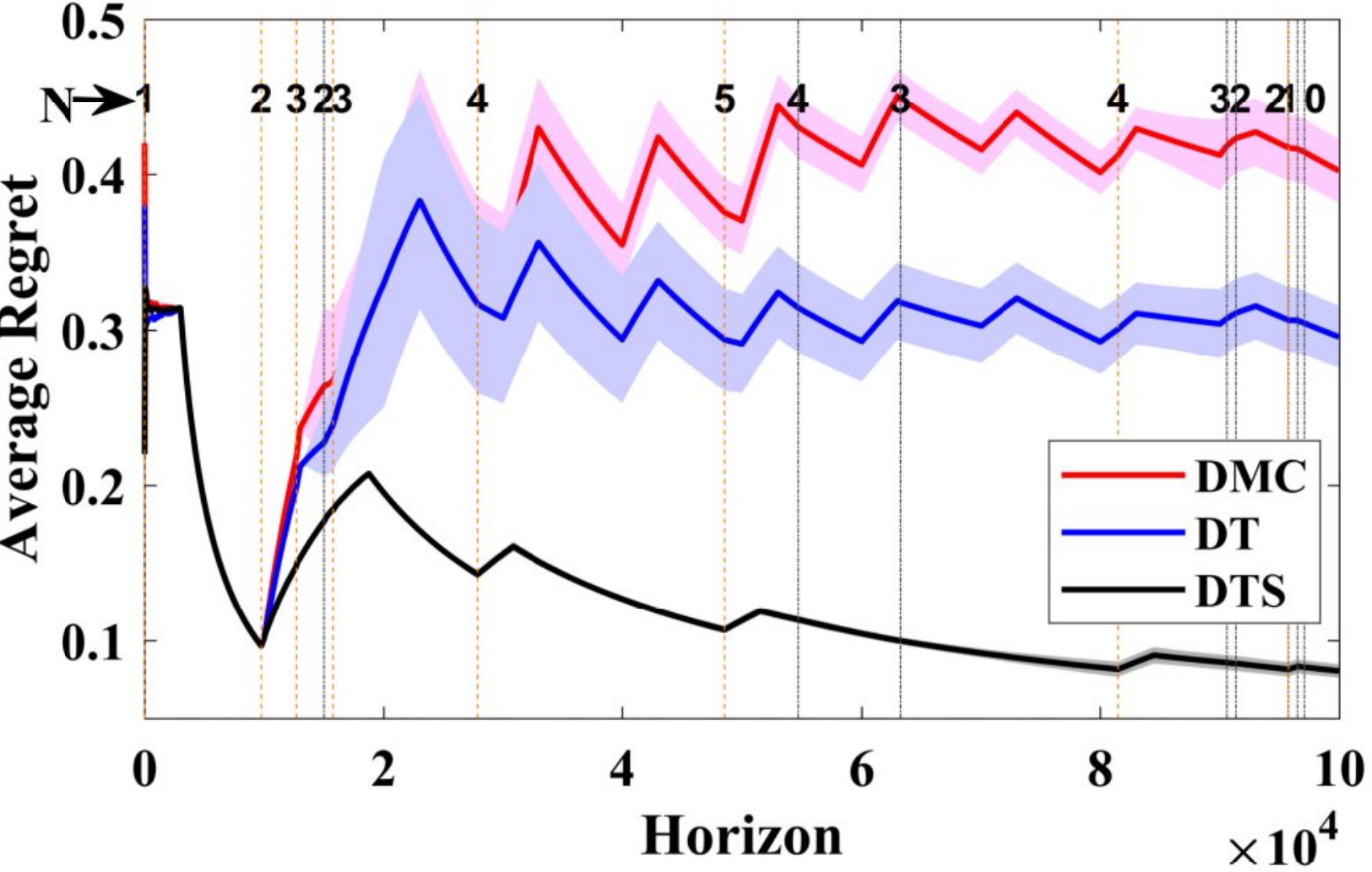}}
	\caption{(a) Cumulative regret and (b) Average regret comparison for Scenario 4 where players can enter or leave any time including the learning phase of the DMC and DT algorithm.}
	\label{dc3}
\end{figure}

\begin{figure}[!h]
	\centering     
	\subfigure[]{\label{dc41}\includegraphics[width=40mm]{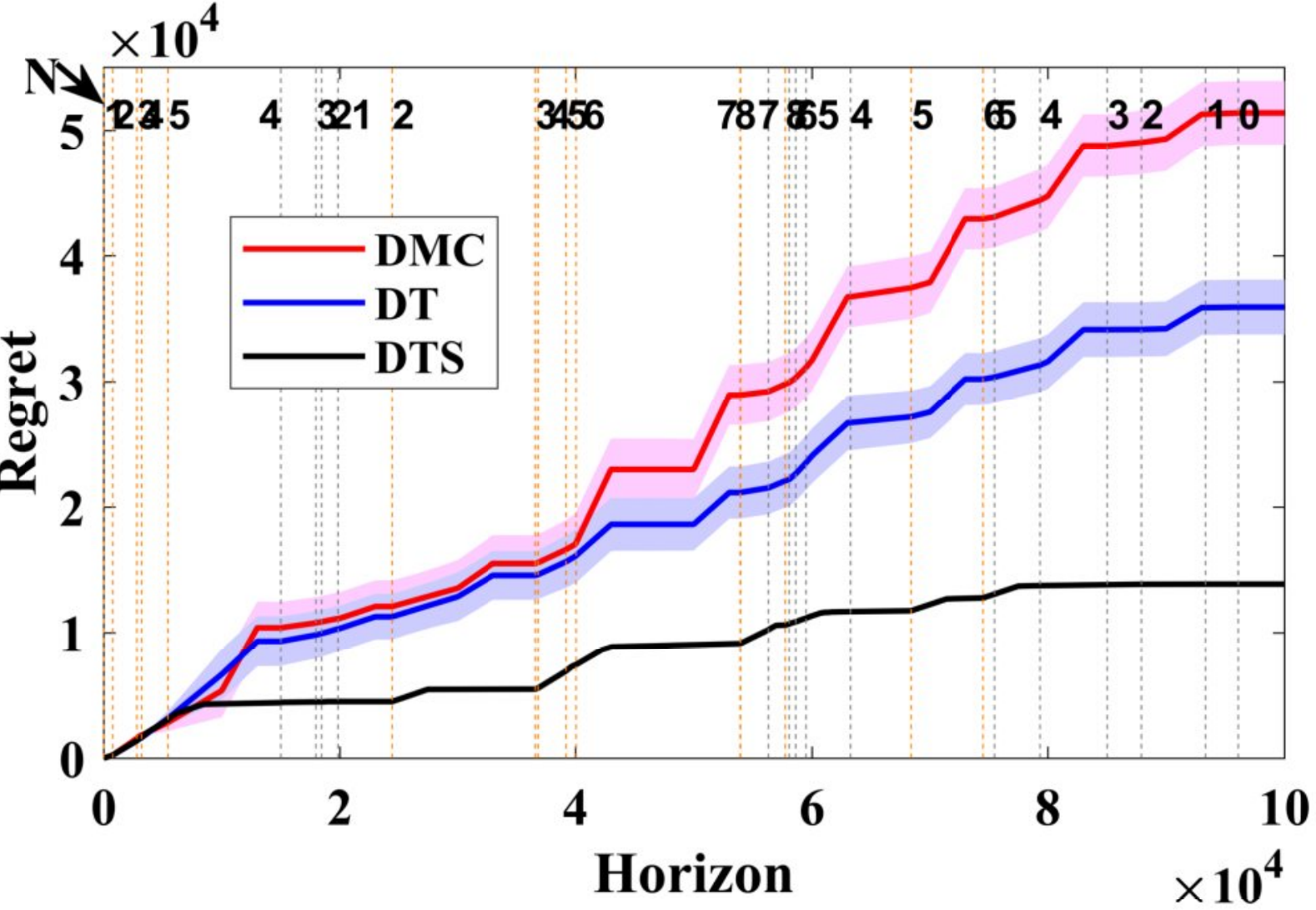}}  \hspace{0.25cm}
	\subfigure[]{\label{dc42}\includegraphics[width=41mm]{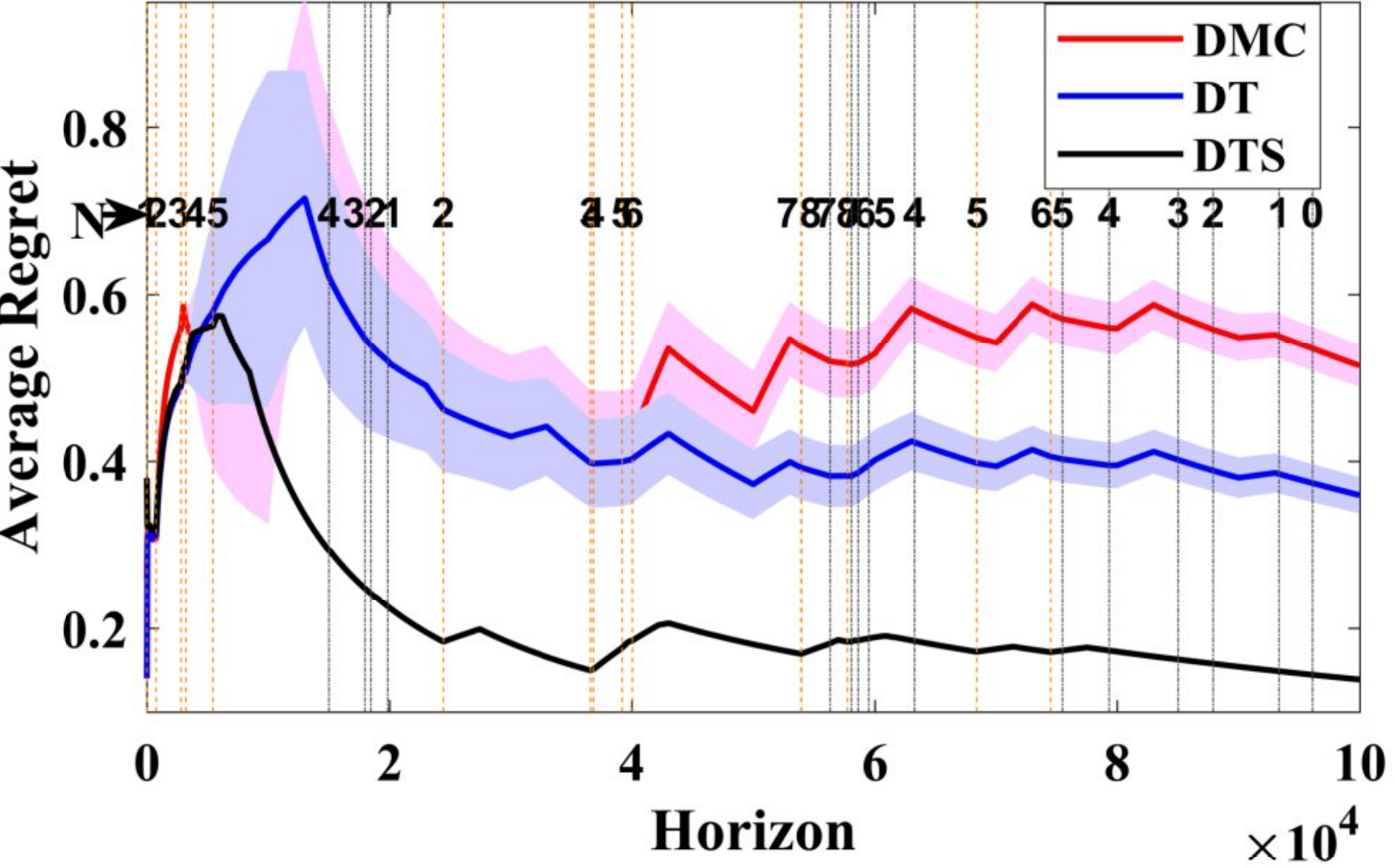}}
	\caption{(a) Cumulative regret and (b) Average regret comparison for Scenario 5 where players can enter or leave any time including the learning phase of the DMC and DT algorithm.}
	\label{dc4}
\end{figure}

In this sub-section, we allow players to enter or exit the game at any round. The DMC algorithm do not allow the player to enter or leave the game during learning and MC phases while DT algorithm restricts entry during learning and trekking phases. In case of DTS algorithm, there is no such restriction on the player entry or exit during the game. It can be observed from the Fig.~\ref{dc3} and Fig.~\ref{dc4} that DTS algorithm performs significantly better than the DT and DMC algorithms. Though the DT algorithm is superior to the DMC algorithm, the difference between the regret of the DT and DMC algorithm is smaller than the scenarios where the players can not enter or exit during the learning and trekking/MC phases.

\begin{figure}[!h]
	\centering     
	\subfigure[]{\label{dc01}\includegraphics[width=40mm]{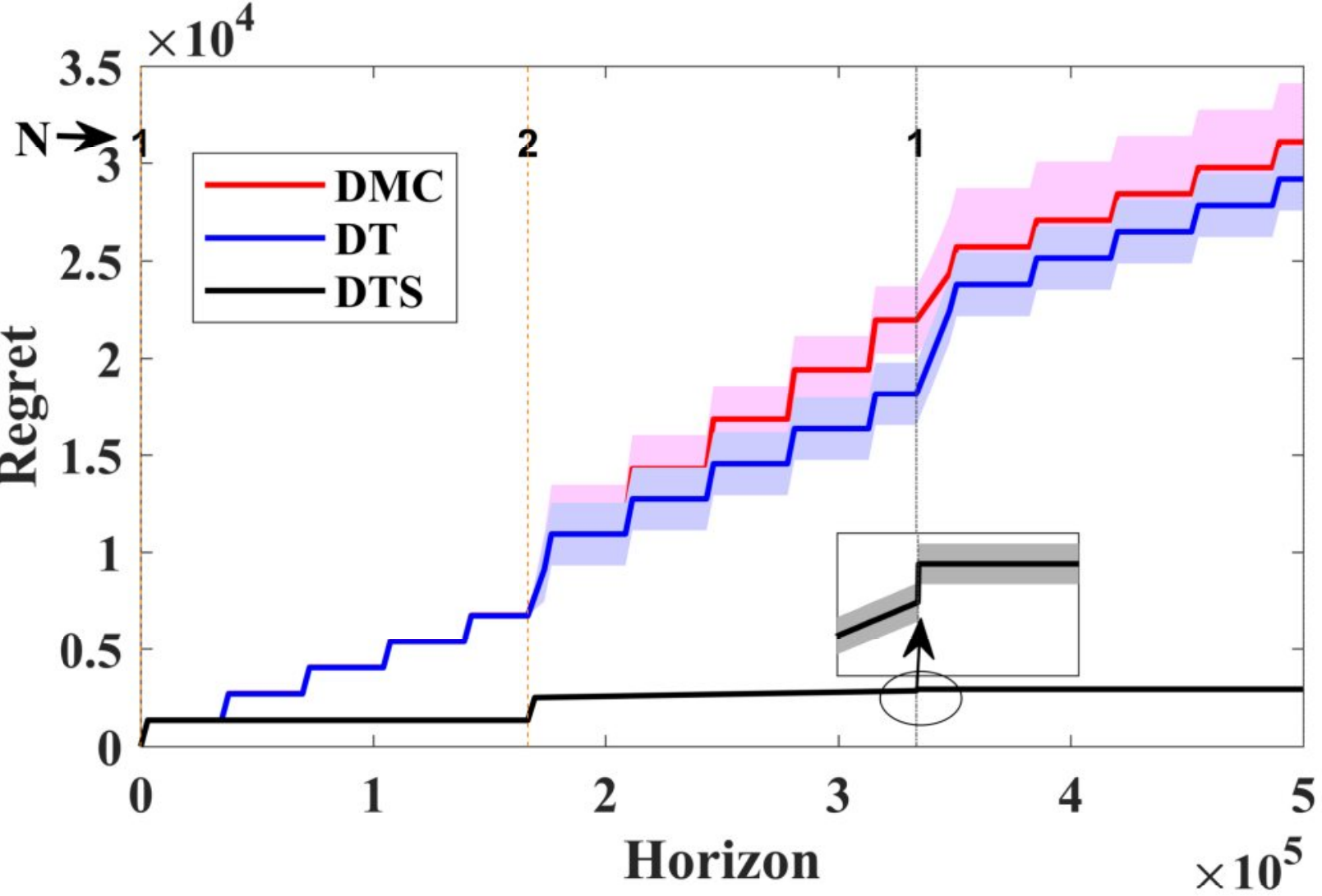}}  \hspace{0.25cm}
	\subfigure[]{\label{dc02}\includegraphics[width=41mm]{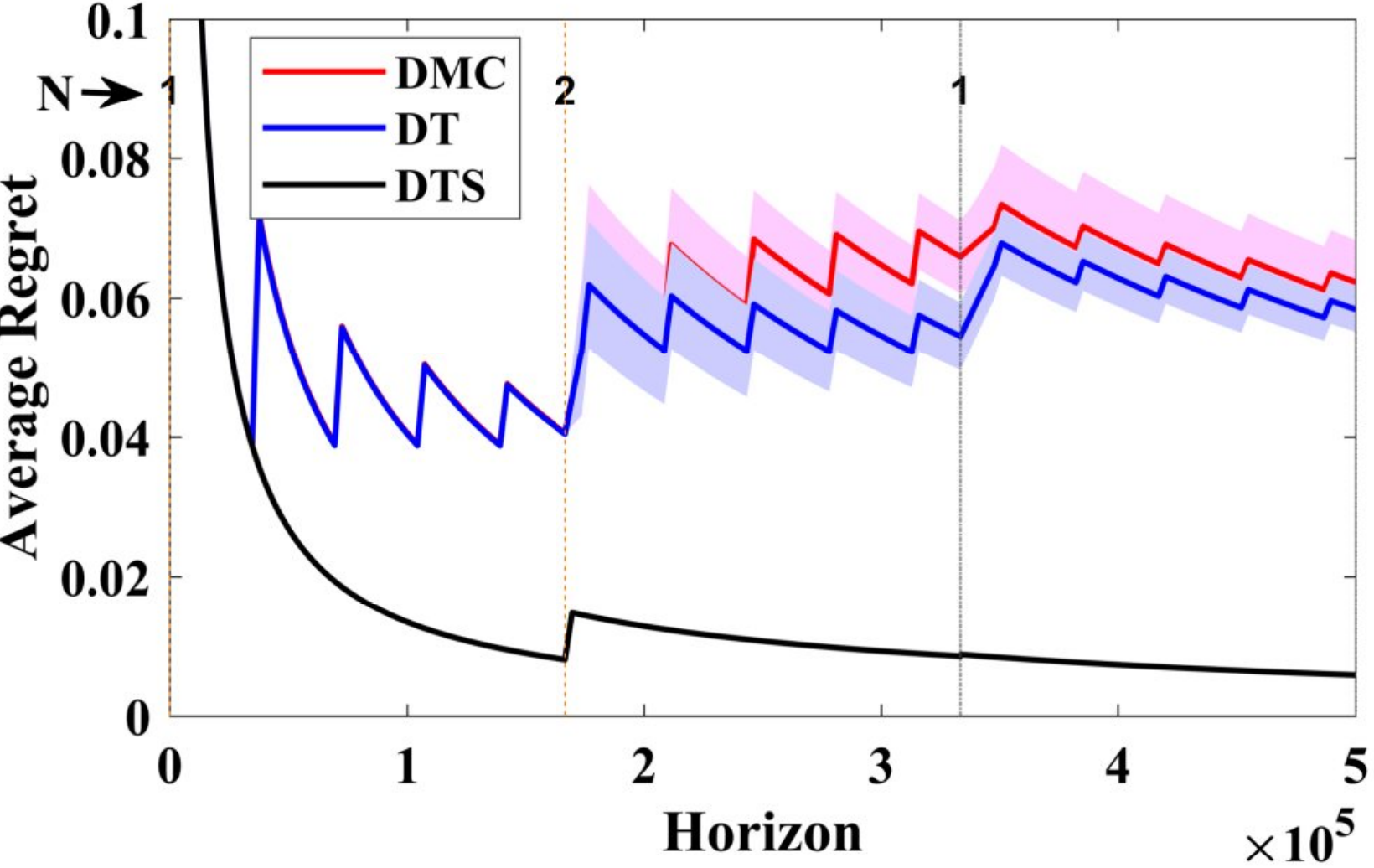}}
	\caption{(a) Cumulative regret and (b) Average regret comparison for Scenario 1 where players can enter or leave any time except during learning phase of the DMC and DT algorithm.}
	\label{dc0}
\end{figure}

\begin{figure}[!h]
	\includegraphics[scale=0.4]{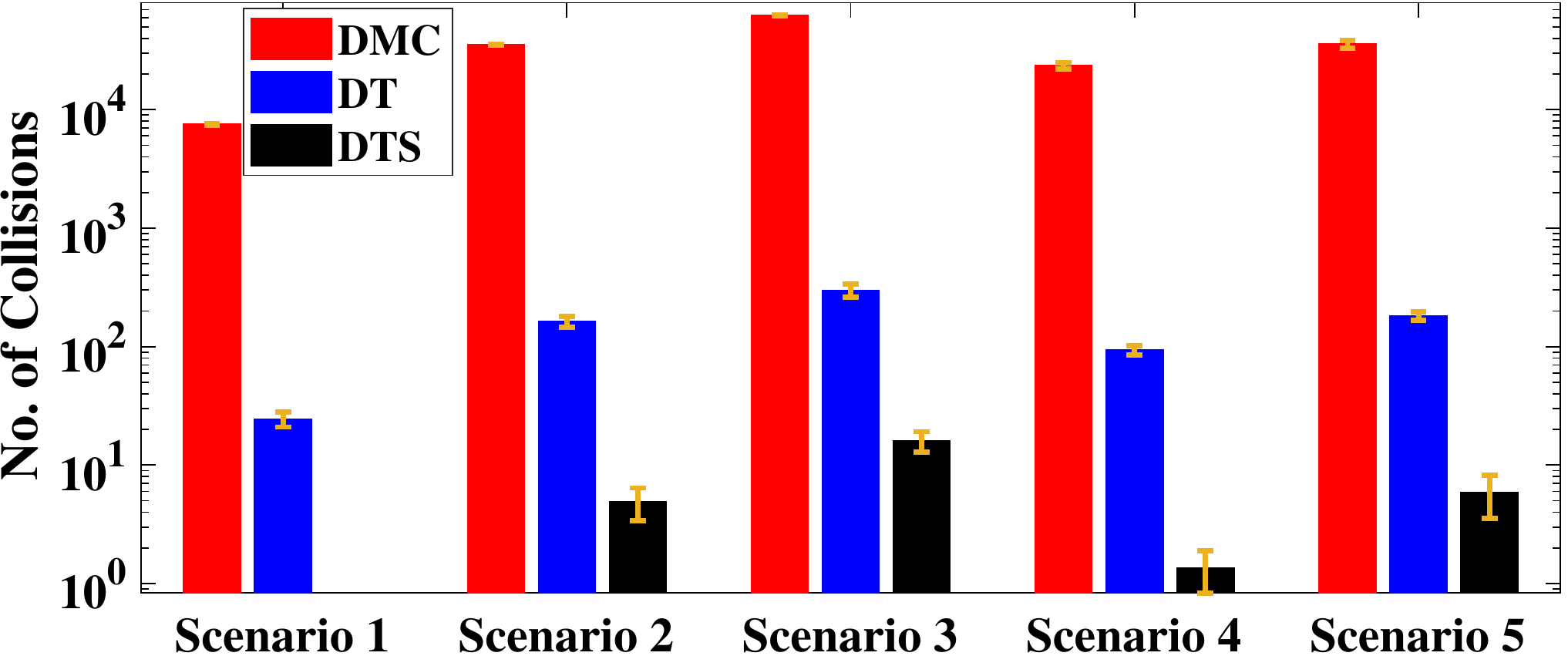}%
	\centering
	\caption{Collision comparison for different scenarios considered for dynamic case with logarithmic scale on $y-axis$.}
	\label{dcl}
\end{figure}

For the five different scenarios considered for dynamic case, it can be observed that the difference between the regret as well as number of collisions of the DMC algorithm and proposed algorithms increases as the number of players and the entering/leaving rate is increased. Thus, higher the dynamism of the game/network, better is the performance of the proposed algorithm compared to the state-of-the-art DMC algorithm. Similar observation is also valid for the number collisions. The Fig.~\ref{dcl} shows that the proposed algorithms offer a significantly fewer number of collisions than the DMC algorithm for all the scenarios considered in dynamic case. The difference between the number of collisions in the DMC algorithm and the proposed algorithms increases significantly as the number of players and the entering/leaving rate increases.

\subsubsection{Scenario 6-7: Special cases}
\begin{figure}[!h]
	\centering     
	\subfigure[]{\label{dt11}\includegraphics[width=40mm]{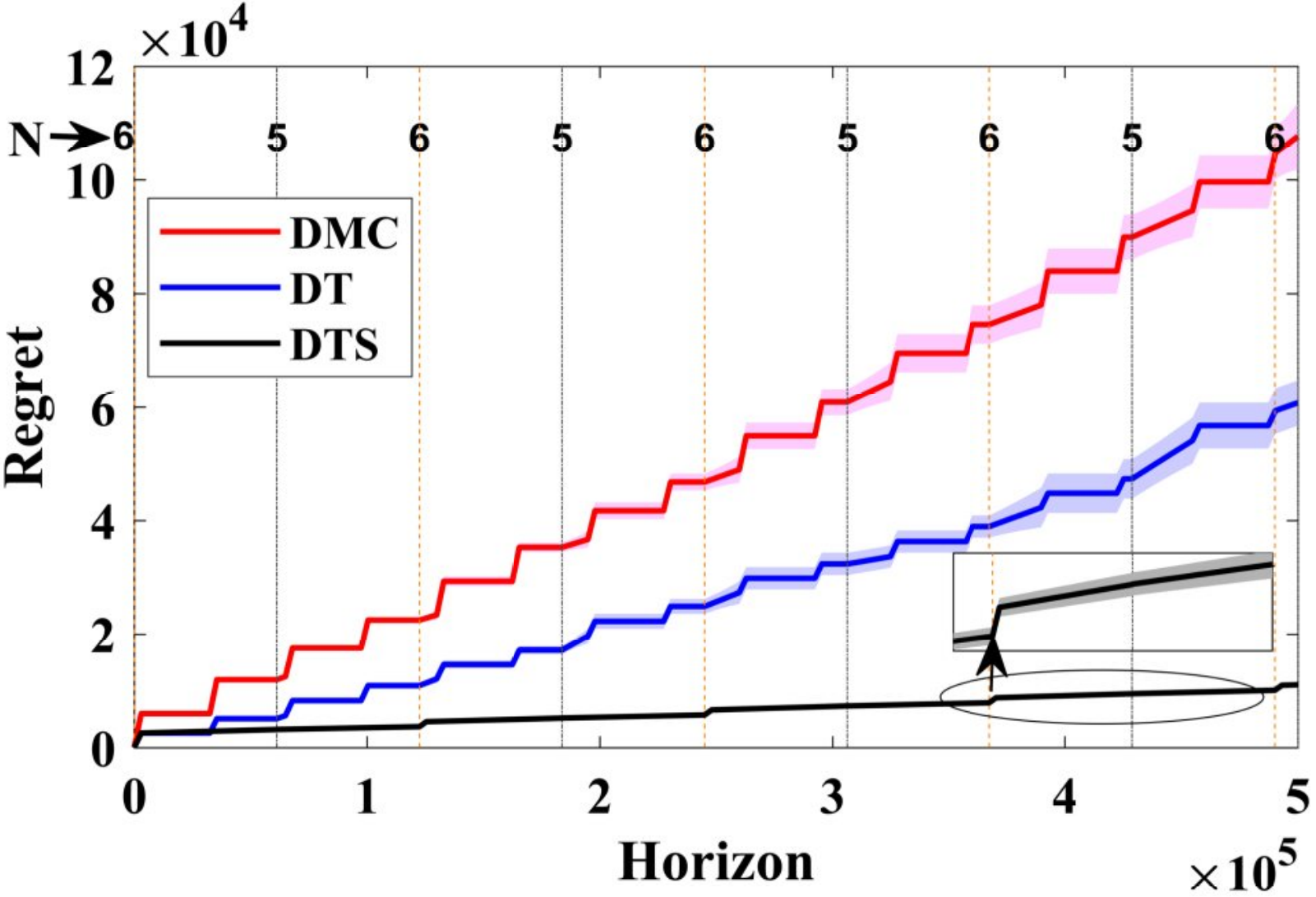}} 
	\subfigure[]{\label{dt12}\includegraphics[width=40mm]{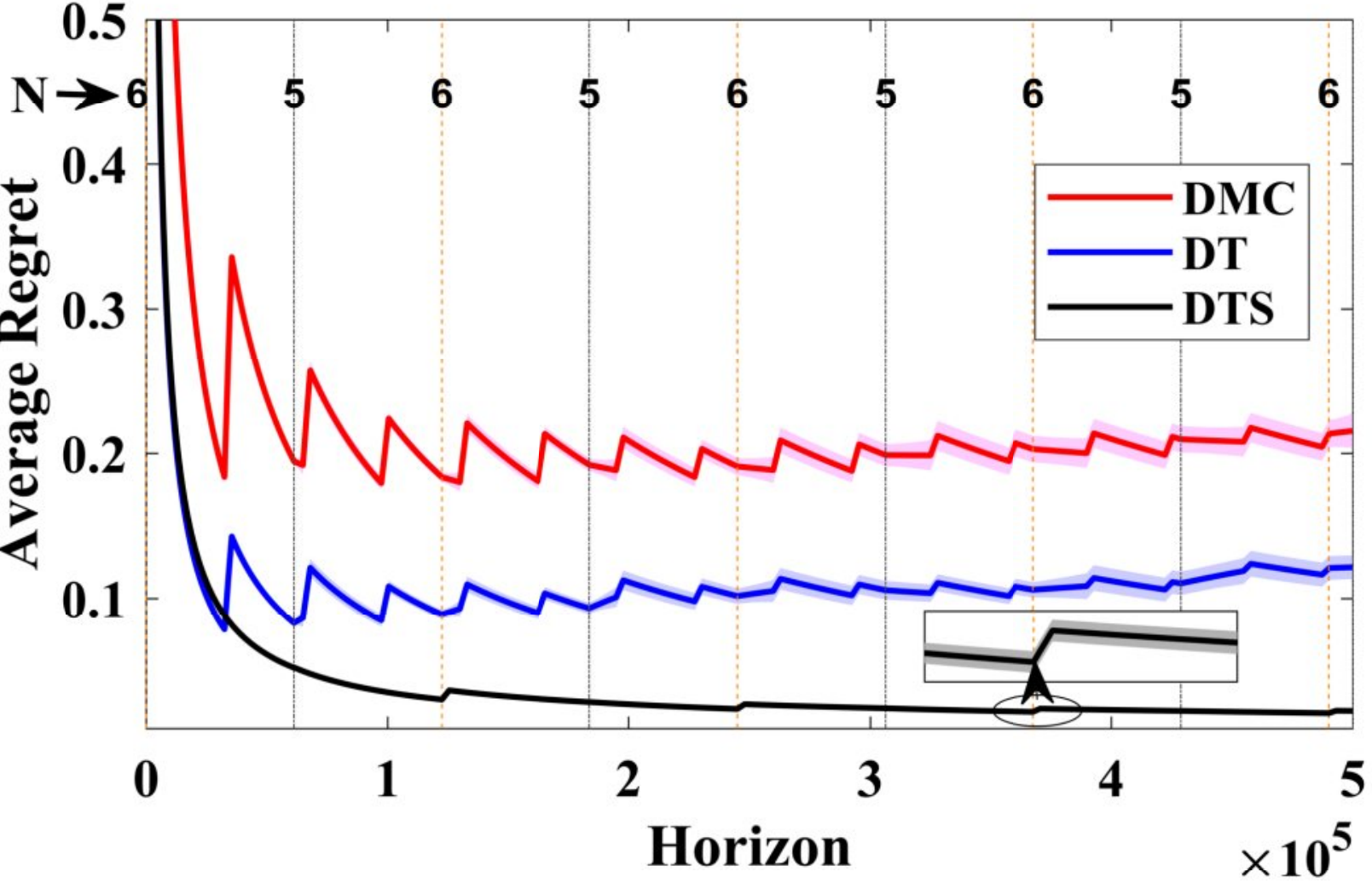}}
	\caption{(a) Cumulative regret and (b) Average regret comparison for dynamic case for Scenario 6.}
	\label{dt1}
\end{figure}

\begin{figure}[!h]
	\centering     
	\subfigure[]{\label{dt21}\includegraphics[width=40mm]{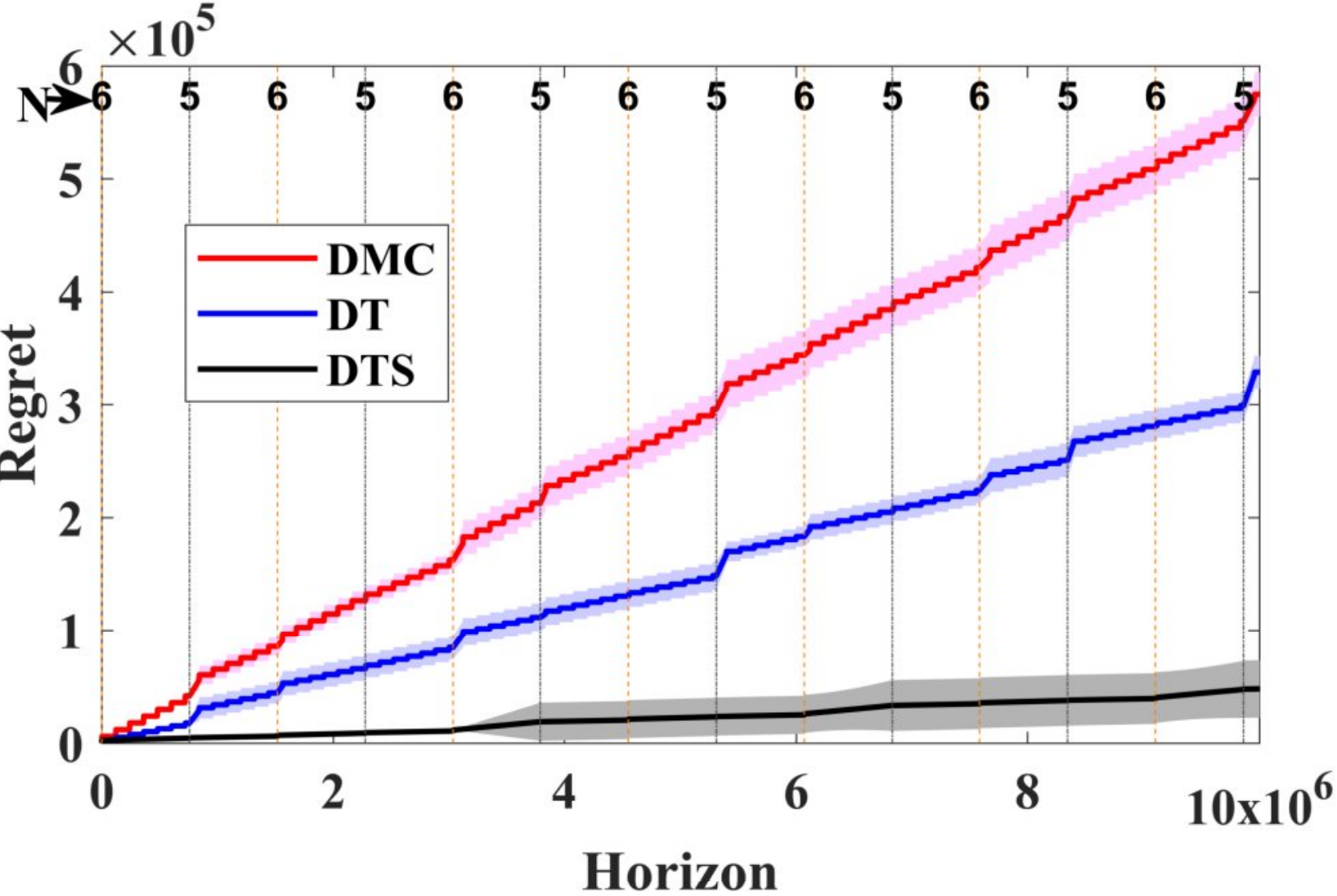}} 
	\subfigure[]{\label{dt22}\includegraphics[width=41mm]{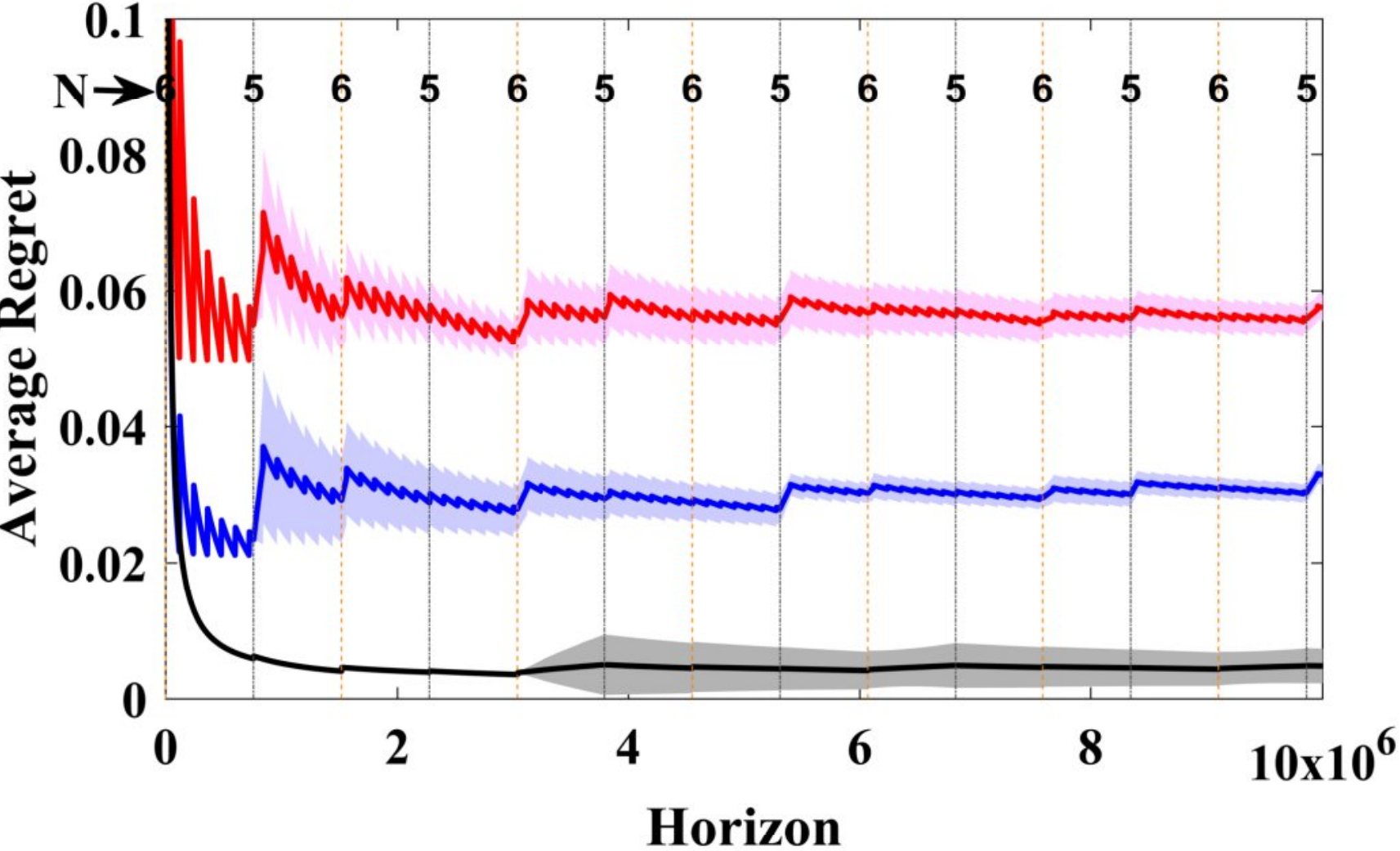}}
	\vspace{-0.45cm}
	\caption{(a) Cumulative regret and (b) Average regret comparison for dynamic case for Scenario 7.}
	\label{dt2}
\end{figure}

Next two scenarios are same as the one considered in \cite{ICML16_MultiplayerBandits_RosenkiShamir}. We begin with the Scenario $6$ where the game starts with a set of six players and $10$ arms with statistics, $\boldmath{\mu}^1$. At every $T^{0.84}$ rounds, we alternate between a player leaving and a player entering the game. The leaving player is chosen at random from the set of current players. Fig.~\ref{dt11} and Fig.~\ref{dt12} show the cumulative regret and average regret, respectively, for the DMC, DT and DTS algorithms with $T= 5*10^5$. We also plot the results for larger horizon of  $T= 10*10^6$ rounds (Scenario $7$). Figure~\ref{dt21} and Figure~\ref{dt22} show the cumulative regret and average regret, respectively, for the DMC, DT and DTS algorithms for long horizon of $T= 10*10^6$ rounds. It can be observed that proposed DT and DTS algorithms offer better performance than the DMC algorithm for small as well as large horizons. 


\subsubsection{Scenario 2: Testbed}
\begin{figure}[!h]
	\includegraphics[scale=0.4]{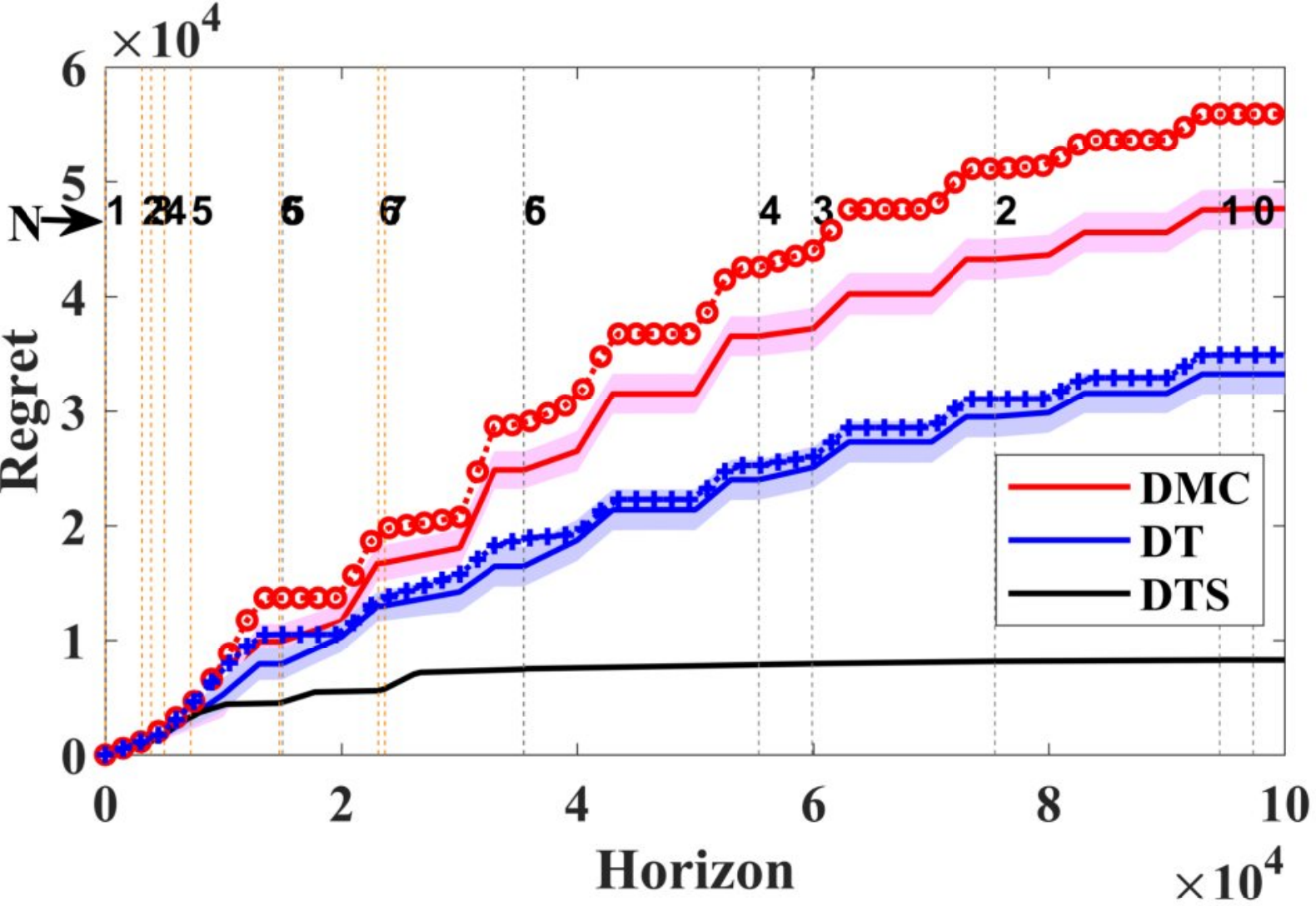}%
	\centering
\caption{Cumulative regret comparison for dynamic case for Scenario 2. The plots with markers refer to experimental results using an actual CR-Network testbed.}
	\label{dt3}
\end{figure}

In Fig.~\ref{dt3}, we include results of the experiments conducted in real radio environment (shown in dotted lines) for the DMC and DT algorithms using the CRN set-up with network parameters set exactly as in Scenario $2$ . The simulation and experimental results show identical behavior validating the feasibility of the proposed algorithm in real environment. Similar results were observed for other scenarios as well but we omitted them due to limited space constraints. All the simulation and experimental results validate the analytical guarantees and gains of proposed algorithms over existing algorithms.


\section{Conclusions and Future Directions}
\label{sec:Conclu}

In this work we introduced new algorithms for stochastic multi-player multi-armed bandits which achieve good regret performance with fewer collisions. The algorithms are completely decentralized and communication-free, and are based on trekking approach where players continuously look for a better arm and occupy it when available. For the static case (fixed number of players) we proposed Static Trekking that improves regret performance by a factor of $4$ and reduces number of collision by a significant factor of $12.K$ over the state-of-the-art algorithm. For the dynamic case (varying number of players) we proposed Dynamic Trekking (DT) that carried over the gains achieved in static setting to the dynamic setting.  For the dynamic setting, we proposed epoch-free Dynamic Trekking with Sensing (DTS) that eliminates the need to have global clock for synchronization and allows players to enter and leave the game anytime.

In this case we considered that the quality of the arms is the same across all the players. In future we would like to study the setting where quality of arms could potentially differ across the players and aim to develop completely distributed algorithms as done in this work.  Also, our algorithms required the knowledge of time horizon $(T)$ to achieve sub-linear regret. It is interesting to see if it is possible to achieve sub-linear regret without knowledge of $T$, especially in the dynamic case.

%
%

\appendices
\label{Appendix}
\subsection*{Proof of Thm \ref{thm:StaticST}}
\label{APA}
We prove Theorem $1$ using the following lemmas which gives expected number of rounds for players to 1) orthogonalize through random hopping 2) learn $\epsilon$-correct ranking of arms and 3) settle on the top $N$ arms. The first two events correspond to  random selection and sequential selection of arms in the learning phase, respectively.  We refer to them as random hopping (RH) and sequential hopping (SH) sub-phase.

\begin{lemma}
	\label{lma:RanHop_Static}
	Let $\delta \in (0,1)$. If RH sub-phase is run for  $ T_{rh}(\delta):=\left \lceil \frac{\log(\delta/K)}{ \log\left(1-1/4K\right)} \right \rceil $ number of rounds then all the players will orthogonalize with probability at least $1-\delta$.  
\end{lemma}  

\noindent \textbf{Proof:}  Let $p_c$ denote the collision
probability of a players when all the players are randomly selecting an arm to play from $[K]$ in each round $t$. Probability that each a player  will observe a collision-free play on an arm after  $T_{01}$ rounds is given by:
\[\sum_{t=1}^{T_{01}}p_c^{t-1}(1-p_c).\]
Setting this value to be at least larger than $1- \frac{\delta}{N}$ for each player we get
\begin{eqnarray}
	\lefteqn{\sum_{t=1}^{T_{01}} p_c^{t-1}(1-p_c) \geq 1- \frac{\delta}{N}} \nonumber\\
	&\iff&1-{p_c}^{T_{01}} \geq 1- \frac{\delta}{N} \nonumber \\
	&\iff& T_{01} \log{p_c} \leq \log\bigg({\frac{\delta}{N}}\bigg) \nonumber \\
	&\iff& T_{01} \geq \frac{\log\big({\frac{\delta}{N}}\big)}{\log{p_c}} \label{eqn:TRHBound}.
\end{eqnarray}
We next give an uniform upper bound on $p_c$. Note that in any round some players may be selecting arms sequentially (call them SH players) while others uniformly at random (call them RH players). Fix a round $t$ and let $N_r \geq 1 $ denote the number of players selecting arms uniformly at random. Let $p_{cr}$ denote the probability that collision is observed from a RH player. We have

\begin{eqnarray*}
\lefteqn{1-p_c= \Pr\{\mbox{no colision from RH players}\}} \\
&+& \Pr\{\mbox{no colision from SH players}\} \\
&\geq & \sum_{j=1}^{N_r}\frac{(1-p_{cr})}{K}\geq \frac{(1-p_{cr})}{K} \\
&=& \frac{(1-1/K)^{N_r-1}}{K}\\
&\geq& \frac{(1-1/K)^{N-1}}{K}\geq 1/4K (\mbox{ for all } K>1).
\end{eqnarray*}  

Substituting the bound on $p_c$ in (\ref{eqn:TRHBound}) and using union bound we see that within
$ T_{rh}(\delta)$ rounds all the players will orthogonalize with probability at least $1-\delta$. \hfill\IEEEQED


%

\begin{lemma}
	\label{lma:Learning_Static}
	For any $\epsilon>0$ and $\delta \in (0,1)$ let $T_{sh}(\delta):=\frac{2K}{\epsilon^2}\log(2KN/\delta)$. If the learning phase is run for $T_0(\delta) := T_{sh}(\delta/2) + T_{rh}(\delta/2)$ rounds, then all the players will have $\epsilon$-correct ranking of arms with probability atleast $1-\delta$.
\end{lemma}
{\bf Proof:} The proof of this lemma is similar to Lemma $1$ in \cite{Arxiv15_MultiplayerBandits_RosenkiShamir}. We repeat it here for completeness. From Lemma \ref{lma:RanHop_Static}, after $T_{rh}(\delta/2)$ rounds of the learning phase (RH sub-phase) all the players are orthogonalized with probability at least $\delta/2$. Conditioned on this event, we will show that players learn $\epsilon$-ranking of arms with probability atleast $\delta/2$ after $\frac{2K}{\epsilon^2}\log(4KN/\delta)$ number of rounds.

Recall a player has an $\epsilon$-correct rank of arms if $\forall k \in [K]$ she has an estimate $\hat{\mu}_j$ such that $\left | \hat{\mu}_k-\mu_k \right| \leq\frac{\epsilon}{2} $. We will upper bound the probability that no SU has $\epsilon-$ correct ranking given that each player has $C_{m}$ observations of each arm. 
Consider the following events:

\noindent$O_n$ - event that player $n$ has observed each arm atleast $C_{m}$ number of times.\\
$A$ - event that all players have $\epsilon$-correct ranking. \\
$A_n$ - event that player $n$ has $\epsilon$- correct ranking. \\
$B$ - event that all players have atleast $C_{m}$ observations of each arm. \\
$B_n$ - event that player $n$ has atleast $C_{m}$ observations of each arm.

In the following we use $\overline{X}$ which denotes complement of event $X$.\\

We have
\begin{align*}
&Pr(\overline{A}_n|B_n) \\
&\leq Pr\left ( \exists k \in [K] \mbox{ such that }| \hat{\mu}_k-\mu_k | >\frac{\epsilon}{2} | \ B_n\right) \\
 &\leq \sum_{k=1}^{K} Pr\left ( | \hat{\mu}_k-\mu_k | >\frac{\epsilon}{2} | \ B_n\right) (\mbox{Union bound})\\ 
&= \sum_{k=1}^{K} \sum_{j=C_{m}}^{\infty} Pr\left (| \hat{\mu}_k-\mu_k | >\frac{\epsilon}{2} | \ O_n = j \right)Pr \left( O_n = j| \ B_n \right) \\ 
\tag {By Hoeffding's Inequality} &\leq \sum_{k=1}^{K} \sum_{j=C_m}^{\infty} 2  \exp \left(\frac{-j \epsilon^2}{2} \right) Pr \left( O_n = j| \ B_n \right) \\ 
&\leq \sum_{k=1}^{K} 2 \exp \left(\frac{-C_{m}  \epsilon^2}{2} \right) \sum_{j=C_m}^{\infty} Pr \left( O_n = j| \ B_n \right) \\ 
&\leq \sum_{k=1}^{K} 2 \exp \left(\frac{-C_{m} \cdot \epsilon^2}{2} \right)  \\
&\leq  2K  \exp \left(\frac{-C_m \epsilon^2}{2} \right)
\end{align*}

We can apply Hoeffding's Inequality since each observation of the arm is independent of the number of times we observe that arm. Setting the bound to be  less than $\frac{\delta}{2N}$, we get 
\[   C_{m} \geq  \frac{2}{\epsilon^2} \ln \left ( \frac{4 KN}{\delta} \right) \]

After the RH sub-phase,  the players select orthogonal arms hence there will be no collision and get reward sample in each round. Further, since they select the arms sequentially, the number of plays of  each arm is in the same proposition. Hence if $T_{sh}(\delta/2)=(2K/ \epsilon^2) \ln \left ( 4 KN/ \delta) \right) $ rounds are played after the RH sub-phase, all players will have $\epsilon$-correct ranking of the arms with probability at least $1-\delta/2$ (by applying union bound). \hfill\IEEEQED

\begin{lemma}
\label{lma:Trekking_Static}
In the trekking phase of ST all the players settle on the top $N$ arms in at most $T_{tr}:=(K^2-(N-1)^2)/2 + 1$ number of rounds.
\end{lemma}
{\bf Proof:} Recall that in the trekking phase each player plays its next best arm, say $i>1$,  for at least $(i-1)$ rounds before taking it as their reserved arm. If there are $n$ players with reserved arms better than than arm $i$,  then a player with reserved arm $i$ can lock only on the $(n+1)$th arm and also all other $n$ players lock on the top arms before her. Hence the maximum number of rounds before the player locks on the $(n+1)$th arm is given by
 \[(i-1)+(i-2)\cdots +( n) + 1= \left (\sum_{k=1}^{i-n} i-k\right ) + 1\]
$1$ is added in the summation to count the round in which the player falls-back on its reserved arm and locks.
The worst case happens when one of the player starts the trekking phase with the worst arm as her reserved arm. Thus setting $i=K$ and $n=N-1$ in the above summation we get the maximum number of rounds in the trekking phase as
\begin{eqnarray*}
T_{tr}&=& \left (\sum_{k=1}^{K-(N-1)} K-k\right ) + 1\\
 &\leq& (K^2 - (N-1)^2)/2 + 1. 
\end{eqnarray*}

\hfill\IEEEQED

\noindent {\bf Proof of Thm 1:} From Lemma \ref{lma:Learning_Static} and Lemma \ref{lma:Trekking_Static} all the players settle on the top $N$ arms without any overlap after $T_{rh}(\delta/2)+T_{sh}(\delta/2) + T_{tr}$ rounds with probability at least $1-\delta$ and regret from the subsequent rounds is zero. 
Hence expected regret of ST with probability at least $(1-\delta)$  is
 \[R \leq N(T_{rh}(\delta/2)+T_{sh}(\delta/2) + T_{tr}).\]
 
The upper bound can be tightened as follows. Notice that in the SH sub-phase, each player selects each arm $1/K$ fraction of the time and in particular the top $N$-arms $N/K$ fraction of the time. When a player selects any of the top $N$ arms, her contribution to regret in that round is zero. Hence each player in the SH sub-phase contribute to regret only $(1-N/K)$ fraction of the time. Adding this factor in the above regret bound we get 
 \[R \leq N(T_{rh}(\delta/2)+T_{sh}(\delta/2)(1-N/K) + T_{tr}).\]

We next bound the number of collisions. During the learning phase, collision occurs only in the RH sub-phase. During the trekking phase, each player can experience at most two collisions -- one before and after locking on an arm. For a player collision can happen before locking when she selects an arm on which another player is locked. Collision can happen after locking when another trekking player selects the arm on which she is locked. Thus total number of collision with probability atleast  $1-\delta$ is bounded as 

\[C \leq N T_{rh}(\delta/2) + 2(2N).\]  \hfill\IEEEQED

\subsection*{Length of Modified Trekking Phase:}

\begin{lemma}
	\label{lma:Trekking_Static1}
	In the modified trekking phase (TrekD) presented in Section~\ref{sec:mtrek}, all the players settle on the top $N$ arms in at most $T_{tr}:=(N-1)(K-1)+1$ number of rounds.
\end{lemma}
{\bf Proof:} Recall that in the modified trekking phase each player plays an arm for at most ($K-i+1$) rounds before locking on it. The time taken by any player to get locked is then at most the number of arms tried before locking multiplied by her back-off time. Note that player on arm $2$ (if any) is the last to back-off from every arm that is taken over by another player and is the last to lock. 
Since the player on arm $2$ has to try at most $N-1$ different arms before she locks, she will lock (and so are others) after at most $T_{tr}:=(N-1)(K-1)+1$ rounds.\hfill\IEEEQED

\subsection*{Proof of Thm  	\ref{thm:MCBound}}
The bound on the regret is given in \cite{ICML16_MultiplayerBandits_RosenkiShamir}[Thm 1]. In the MC algorithm the learning phase is run for $T_0^{MC}$ number of rounds in which each player select arm randomly from $[K]$ in each round. The probability of observing a collision for a player in each round is $(1-(1-1/K)^{N-1})$. Hence expected number of collisions are at least

\[C^{MC} \geq NT_0^{MC} (1-(1-1/K)^{N-1}).\]
For $N\geq 2$, we have $(1-1/K)^{N-1} \leq 1-1/K$. Hence we get $C^{MC} \geq NT_0^{MC}/K$ as claimed. \hfill\IEEEQED

\subsection*{Proof of Proposition \ref{prop:DynamicDT}}
The proof of this Theorem follows along the ideas similar to that in \cite{ICML16_MultiplayerBandits_RosenkiShamir}[Thm 2.]

We first bound the regret. The regret in each epoch is composed of the following three terms:

\begin{itemize}
	\item Regret due to learning and trekking phase
	\item Regret due to entering players
	\item Regret due to leaving players 
\end{itemize}
Let $T_0:=T_0(\delta/T)$ denote the length of learning phase in each epoch.
Then with probability $1-\delta/T$ all the players in that epoch will have $\epsilon$-correct ranking of the arms leading to zero regret after the trekking phase.
Note that $T_0$ is a function of $T$ and grows logarithmically in $T$.
 
 \noindent
{\bf Regret due to learning and trekking phase:} The length of the this period is $T_0 + T_{tr}$ and adds at most $K(T_0+ T_{tr})$ regret.

\noindent
{\bf Regret due to entering players:}
Recall that we allow a new player to enter after the learning and trekking phase in each epoch. Each new player collides with at most one player in each round. If $e_i$ is the number of player that enter in an epoch they add 
at most $2e_i(T_{ep}-T_0-T_{tr})$ regret. A factor $2$ is because reward from two optimal arms is lost each time a collision happens.

\noindent
{\bf Regret due to leaving :}
Recall that player can leave at any time. A player leaving during the learning phase do not cause any regret, whereas if a player leaves after the learning phase, the arm on which she was locked may not be taken over by any other player and regret is incurred for the remaining rounds. Hence, if $l_i$ players leave in an epoch, it add at most $l_i (T_{ep}- T_0-T_{tr})$ regret.

\noindent
Let
$e=\sum_ {i=1}^{T/T_{ep}} e_i$ and $l=\sum_ {i=1}^{T/T_{ep}}l_i$ denote the total number of entering and leaving players across all epochs. Combining regret from all the three parts from each epoch and adding over all the epochs, we get
\begin{align*}
R &\leq  \frac{T}{T_{ep}}(K(T_0+ T_{tr})) + 2e (T_{ep} -T_0 -T_{tr}) \\
& \quad + l (T_{ep} -T_0 -T_{tr})
\end{align*}
We set the value of $T_{ep}$ as
\[T_{ep} =\sqrt{\frac{KT(T_0+T_{tr})}{2x}}.\]
which minimizes the upper bound. Finally, the result follows by taking union bound over $T$.

We next bound the number of collisions. Note that a leaving players will not cause any collision. In each epoch number of collision in the learning phase is at most $T_{rh}:=T_{rh}(\delta/2T)$ and the trekking phase is at most $4K$. Each entering player will cause at most $2(T_{ep}-T_0-T_{tr})$. Hence total number of collisions is at most

\[C\leq \frac{T}{T_{ep}}(KT_{rh}+4K)+ 2x(T_{ep}-T_0-T_{tr}).\] \hfill\IEEEQED

\subsection*{Proof of Thm \label{thm:DynamicDT}}
\noindent
From the proof of Lemma $1$, recall that $T_0= \mathcal{O}(\log T)$. Hence
$T_{ep}=\mathcal{O}(\sqrt{T\log T /x})$. We get
\begin{eqnarray*}
R &\leq& \mathcal{O}\left (\frac{T}{T_{ep}}T_0 + xT_{ep} \right ) \\
&=& \mathcal{O} \left( 
\frac{\sqrt{x}T}{\sqrt{T\log T}} \log T +  x \sqrt{\frac{T \log T}{x}}\right) \\
&=& \mathcal{O} \left( 
\sqrt{xT\log T}  +   \sqrt{xT \log T}\right)\\
&=&\tilde{\mathcal{O}}(\sqrt{xT}), 
\end{eqnarray*}
 where $\tilde{\mathcal{O}}$ hides logarithmic factor in $T$.
 
 The bound on the number of collisions also follows similarly by noting that
 \begin{equation}
  C \leq \mathcal{O}\left (\frac{T}{T_{ep}}T_0 + xT_{ep} \right ).
 	\end{equation}
 	
\hfill\IEEEQED

\subsection*{Proof of Theorem \ref{thm:DynamicDTS}}
 \noindent
 We prove the Theorem using the following lemma
 
 \begin{lemma}
 	Consider the same setup as in Prop. \ref{prop:DynamicDT}. For any $\delta \in (0,1)$ let 
      length of the learning phase in DTS is set as 
 	\[
 	\tilde{T}_0=\left \lceil \frac{\log(\delta/2(K+x))}{\log\left(1-1/4K\right)}\right \rceil + \frac{2K}{\epsilon^2}\log \left (\frac{4K(K+x)}{\delta}\right),\]
 	Then, expected regret of DTS after $T$ rounds is bounded with probability at least $1-\delta$ as follows:
 	
 	\begin{equation*}
 	R \leq K\tilde{T}_0+ e\tilde{T}_0 + xKT_l + (T/T_l)\tilde{T}_{tr} K 
 	\end{equation*}
 \end{lemma}
where 
\[\tilde{T}_{tr}:= (K/2)^2 + K/2.\]

  Further, expected number of collisions is bounded with probability at least $1-\delta$ as
 \begin{equation*}
C\leq (K+x)\left \lceil \frac{\log(\delta/2(K+x))}{\log\left(1-1/4K\right)}\right \rceil + (T/T_l)\tilde{T}_{tr}K.
 \end{equation*}

{\bf Proof:} We first bound the regret. The regret is composed of the following terms. 
 
 \begin{itemize}
 	\item Regret due to learning phase of players who joined from the start
 	\item Regret due to learning phase of players entering the game late
 	\item Regret due to players leaving the game
 	\item Regret due to continuous trekking of all players
 \end{itemize}
 	
{\bf Regret due to players who joined from the start:}\\
Let $N_m  \leq K$ denote the number of players that join the game at $t=0$ and $\tilde{T}_0$ denote the length of the learning phase. The regret due to learning phase is upper bounded $K\tilde{T}_0$.

{\bf Regret due to learning phase of players entering late:}\\
Let $e$ denote the number of players that enter the game late. We note that the entering players do not disturb already settled players as they sense selected arms before playing. Thus when a new player enters regret is incurred only due to them and not due to already settled players which is upper bounded by $e\tilde{T}_0$. Further, if the at the end of learning phase, the entering players may not have estimates for mean rewards for some arms and their estimates are obtained during the trekking phase by playing them for $T_l$ rounds each. This will cause additional regret bounded by $KT_l$. The worst case happens when a player do not get estimates of any arm. Hence the regret upper bound due to entering players is
$e\tilde{T}_0 + eKT_l$ 

{\bf Regret due to leaving players:}\\
When a player leaves, the freed-up arm will be taken by one of the existing player after at most $T_l$ rounds -- within $T_l$ rounds after the player leaves one of the existing players enters into the trekking state and takes over the free-up arm. Hence $l$ leaving players cause at most $lT_l$ rounds of regret.

{\bf Regret due to continuous trekking of all players:}\\
We first argue that the regret due to the players having estimates of all or only few of the arms can be treating in the same fashion.  Consider a player that does not have estimates of all the arms (entering/late player). This player checks for an arm for which  estimate is not available yet before checking for the best available arm for the set of estimated arms and locks on it whenever it is available to get its estimate.  At most $KT_l$ (this could be spread over multiple trekking cycles) rounds incurred due to this. This factor is already accounted in the regret computed in the second point. Hence we compute regret during the continuous trekking assuming that all players have estimates of all the arms. We refer to number of rounds a players spends in trekking state before he enters into locked state as one trekking cycle.

We next bound length of a trekking cycle. A player with reserved arm $i$ requires at most $(i-1)(K- i +1) + i$ to complete checking of availability of better arms -- the players spends at most $K-i +1$ rounds (back-off) on each of the arms $1,2, i-1$. If none of them is available she locks back on $i$. Addition of $i$ accounts for each return to arm $i$.  Optimizing over the value of $i$, the maximum length of a trekking cycle is given by $\tilde{T}_{tr}:= (K/2)^2 + K/2$.

Over period $T$, number of trekking cycles for each player is at most $T/T_l$ and in each trekking regret is at most $\tilde{T}_{tr}$. Since at most $K$ players can be in the game at any time, regret due to continuous trekking is upper bound by 
           \[(T/T_l)\tilde{T}_{tr}K.\]

Combining all the terms, the regret is upper bound as

\begin{eqnarray}
R &\leq& K\tilde{T}_0+ e\tilde{T}_0 + eKT_l + lT_l + (T/T_l)\tilde{T}_{tr} K  \nonumber \\
&\leq& K\tilde{T}_0+ e\tilde{T}_0 + eKT_l + lKT_l + (T/T_l)\tilde{T}_{tr} K \nonumber \\
\label{eqn:DTS_RegretBound}
&\leq& K\tilde{T}_0+ e\tilde{T}_0 + xKT_l + (T/T_l)\tilde{T}_{tr} K.
\end{eqnarray}

Setting 
\begin{equation*}
\tilde{T}_0=\left \lceil \frac{\log(\delta/2(K+x))}{\log\left(1-1/4K\right)}\right \rceil + \frac{2K}{\epsilon^2}\log \left (\frac{4K(K+x)}{\delta}\right).
\end{equation*}
and using arguments similar to that in Thm 1, and applying union bound over all players (at most $(K+x)$), the regret bound holds with probability $1-\delta$.

To bound the collision, note that the  players incur collisions during RH sub-phase of the learning phase and  whenever it enters into the trekking state during the trekking phase.  Since at most $(N_m + e)$ players enter into RH sub-phase and at most $K$ player in the game at any time we get that collisions are bounded with probability $1-\delta$ by
\begin{equation}
\label{eqn:DTS_Collision}
C\leq (K+x)\left \lceil \frac{\log(\delta/2(K+x))}{\log\left(1-1/4K\right)}\right \rceil + (T/T_l)\tilde{T}_{tr}K.
\end{equation} 
\hfill\IEEEQED

We now return to the proof of Theorem \ref{thm:DynamicDTS}. Ignoring the constants, the regret bound in \ref{eqn:DTS_RegretBound} is given by 
\[R \leq \mathcal{O}(xKT_l + (T/T_l)\tilde{T}_{tr} K )\]
Differentiating the bound w.r.t to $T_l$, we find the optimal value of the bound is given as
\[R \leq \mathcal{O}\left(2K\sqrt{xT\tilde{T}_{tr}}  \right).\]
and it is achieved by setting
\[T_l=\sqrt{\frac{T\tilde{T}_{tr}}{x}}.\]
Similarly, by plugging the above value of $T_l$ in (\ref{eqn:DTS_Collision}) we get
\[C \leq \mathcal{O}\left( K\sqrt{xT\tilde{T}_{tr}} \right).\]
\hfill\IEEEQED

\ifCLASSOPTIONcaptionsoff
\newpage
\fi

\bibliographystyle{IEEEtran}
\bibliography{biblio}

%


\begin{IEEEbiographynophoto}{Manjesh K. Hanawal}
	received the M.S. degree in ECE from the Indian Institute of Science, Bangalore, India, in 2009,
	and the Ph.D. degree from INRIA, Sophia Antipolis, France, and the University of Avignon, Avignon, France, in 2013. After spending two
	years as a postdoctoral associate at Boston University, he is now an Assistant Professor in Industrial Engineering and Operations Research
	at the Indian Institute of Technology Bombay, Mumbai, India. His research interests include communication networks, machine learning
	and network economics.
\end{IEEEbiographynophoto}

\begin{IEEEbiographynophoto}{Sumit J. Darak}
	received his bachelor degree in ECE from Pune University, India in 2007, and PhD degree from the
	School of Computer Engineering, Nanyang Technological University (NTU),
	Singapore in 2013. He is currently an Assistant Professor at Indraprastha
	Institute of Information Technology, Delhi (IIIT-Delhi), India. From March 2013 to November
	2014, he was a postdoctoral researcher at the CentraleSupélec, France.
	Dr. Sumit has been awarded India Government's \textit{DST Inspire Faculty Award}
	which is a prestigious award for young researchers under 32 years age. He has received \textit{Best Demo Award} at CROWNCOM 2016, \textit{Young Scientist Paper Award} at URSI 2014 and 2017, \textit{Best Student Paper Award} at IEEE DASC 2017. His current research interests include the reinforcement learning algorithms and reconfigurable architectures for applications such as wireless communications, energy harvesting etc.
\end{IEEEbiographynophoto}
\vspace{-1cm}

%
%
%

%
%




\end{document}